\documentclass[preprint, 12pt]{elsarticle}

\journal{Neurocomputing}
\usepackage[utf8]{inputenc} 
\usepackage[T1]{fontenc}    
\usepackage{hyperref}       
\usepackage{url}            
\usepackage{booktabs}       
\usepackage{amsfonts}       
\usepackage{nicefrac}       
\usepackage{microtype}      
\usepackage{makeidx}
\usepackage{subcaption}
\usepackage{multirow}
\usepackage{multicol}
\usepackage[dvipsnames,svgnames,table]{xcolor}
\usepackage{graphicx}
\usepackage{float}
\usepackage{epstopdf}
\usepackage{appendix}
\usepackage{amsmath}
\usepackage{cleveref}
\usepackage{amssymb}
\usepackage{csquotes}
\usepackage{color}
\usepackage{mathtools}
\usepackage[ruled,vlined]{algorithm2e}
\usepackage{algorithmic}
\usepackage[normalem]{ulem}
\usepackage{hyperref}

\usepackage{comment}
\usepackage{tabularx}
\usepackage{tikz}
\usetikzlibrary{fit,positioning}

\newcommand{\review}[1]{{\color{black}{#1}}}
\title{A conditional one-output likelihood formulation for multitask Gaussian processes}

\author[1]{\'Oscar Garc\'ia-Hinde \corref{cor1}}
\author[2]{Manel Mart\'{i}nez-Ram\'on}
\author[1]{Vanessa  G\'omez-Verdejo \corref{cor1}}

\cortext[cor1]{Corresponding authors. Email address: oghinde@tsc.uc3m.es, vanessa@tsc.uc3m.es}

\address[1]{Department of Signal Processing and Communications, Universidad Carlos III de Madrid Legan\'es, 28911 Spain
}

\address[2]{Department of Electrical and Computer Engineering, University of New Mexico, Albuquerque, NM, 8711 USA}

\begin{document}
\begin{frontmatter}

\begin{abstract}
Multitask Gaussian processes (MTGP) are the Gaussian process (GP) framework's solution for multioutput regression problems in which the $T$ elements of the regressors cannot be considered conditionally independent given the observations.
Standard MTGP models assume that there exist both a multitask covariance matrix as a function of an intertask matrix, and a noise covariance matrix. These matrices need to be approximated by a low rank simplification of order $P$ in order to reduce the number of parameters to be learnt from $T^2$ to $TP$.
Here we introduce a novel approach that simplifies the multitask learning by reducing it to a set of conditioned univariate GPs without the need for any low rank approximations, therefore completely eliminating the requirement to select an adequate value for hyperparameter $P$.
At the same time, by extending this approach with both a hierarchical and an approximate model, the proposed extensions are capable of recovering the multitask covariance and noise matrices after learning only $2T$ parameters, avoiding the validation of any model hyperparameter and reducing the overall complexity of the model as well as the risk of overfitting. 
Experimental results over synthetic and real problems confirm the advantages of this inference approach in its ability to accurately recover the original noise and signal matrices, as well as the achieved performance improvement in comparison to other state of art MTGP approaches. 
We have also integrated the model with standard GP toolboxes, showing that it is computationally competitive with state of the art options.
\end{abstract}


\end{frontmatter}







\newcommand{\bc}{{\bf c}}
\newcommand{\bx}{{\bf x}}
\newcommand{\by}{{\bf y}}
\newcommand{\bw}{{\bf w}}
\newcommand{\be}{{\bf e}}
\newcommand{\bbf}{{\bf f}}
\newcommand{\bb}{{\bf b}}
\newcommand{\bu}{{\bf u}}
\newcommand{\bA}{{\bf A}}
\newcommand{\bB}{{\bf B}}
\newcommand{\bC}{{\bf C}}
\newcommand{\bD}{{\bf D}}
\newcommand{\bI}{{\bf I}}
\newcommand{\bK}{{\bf K}}
\newcommand{\bk}{{\bf k}}
\newcommand{\bz}{{\bf z}}
\newcommand{\bmu}{{\boldsymbol \mu}}

\newcommand{\bsigma}{{\boldsymbol \sigma}}

\newcommand{\bZero}{{\bf 0}}

\newcommand{\bW}{{\bf W}}
\newcommand{\bX}{{\bf X}}
\newcommand{\bY}{{\bf Y}}
\newcommand{\bZ}{{\bf Z}}
\newcommand{\bF}{{\bf F}}
\newcommand{\bU}{{\bf U}}
\newcommand{\bS}{{\bf S}}
\newcommand{\bV}{{\bf V}}
\newcommand{\bm}{{\bf m}}

\newcommand{\bSigma}{{\boldsymbol \Sigma}}
\newcommand{\balpha}{{\boldsymbol \alpha}}
\newcommand{\btheta}{{\boldsymbol \theta}}
\newcommand{\bLambda}{{\boldsymbol \lambda}}
\newcommand{\bepsilon}{{\boldsymbol \epsilon}}
\newcommand{\bPhi}{{\boldsymbol \Phi}}
\newcommand{\bphi}{{\boldsymbol \phi}}

\newcommand{\manel}[1]{{\color{blue}{#1}}}

%

\section{Introduction}

Gaussian processes (GP) \citep{rasmussen2006} can be considered state of the art in nonlinear regression, among other reasons, because they provide a natural way to implement a predictive posterior distribution. This distribution has a clear advantage over non-Bayesian models since  it is a surrogate statistical model that provides not only the predictive target means, but also a relevant measure of uncertainty in the form of a predictive covariance function. While standard GPs were initially designed to handle single scalar outputs, it is becoming more and more common to have to face multi-task (MT) or multidimensional output problems in which each individual output cannot be considered conditionally independent from the rest given the predictors. 
 This can be found in many application examples, \review{such as medical applications \cite{wiens2016patient, BOUBNOVSKI2022}, air quality forecasting \cite{sun2021deep}, product design \cite{turetskyy2021} and, particularly with multioutput GP models, in multioutput time-series analysis \cite{13},  manufacturing applications \cite{Shen22}, detection of damages in structures \cite{LI2021110085}, forecast of multiscale solar radiation with application in photovoltaics  \cite{ZHOU2021124710}, or COVID-19 outbreak detection \cite{Ketu21}. See also the extensive survey on multi-output learning in \cite{Xu20}}. In these cases, the use of adequate approaches which are able to model the relationships among the tasks can offer significant advantages \cite{adiyeke2020benefits}.


The general MTGP formulation proposed in \cite{Bonilla2008} can be considered the reference model and we will therefore refer to it in this paper as the standard MTGP (Std-MTGP).  \review{Indeed, this model is the one chosen in all the above mentioned  MTGP applications}. This model assumes that the MT covariance matrix is expressed as the Kroneker product of an inter-task matrix $\bC$ and the input kernel matrix. Specifying a full rank $\bC$ requires a computational cost of $\mathcal{O}(N^3T^3)$ and the inference of $T(T- 1)/2$  parameters, which becomes computationally unwieldy when $T$ is large. In order to circumvent these problems, the authors of the Std-MTGP use a low-rank approximation of order $P$ of $\bC$, $\bC\approx \bU \bLambda \bU^\top+\sigma^2\bI$, so that the number of parameters to be learnt is reduced from $\mathcal{O}(T^2)$ to $\mathcal{O}(TP)$ and the computational burden of the method is reduced from $\mathcal{O}(N^3T^3)$ to $\mathcal{O}(N^3T^2P)$. This model is reformulated as a linear model of coregionalization (LMC) in \cite{schmidt2003Bayesian, fanshawe2012bivariate}, where the model outputs are expressed as a linear combination of $P$ latent functions and, therefore, the multitask kernel function can be also expressed as a linear combination of several covariance functions. In both cases, parameter $P$ must be cross validated in order to obtain matrices that are representative of the process to be modelled, but authors usually choose a low value for parameter $P$ in order to keep the model's complexity low. In the particular case where $P=1$, this model simplifies into the well-known intrinsic coregionalization model (ICM) \cite{goovaerts1997geostatistics}, which results in significant computational savings. To obtain further computational savings in the general MTGP formulation, \cite{stegle2011efficient} proposes an efficient inversion of the MT covariance matrix by combining properties of the singular value decomposition (SVD) and the Kronecker product, reducing the computational cost from $\mathcal{O}(N^3T^2P)$ to $\mathcal{O}(N^3+T^3)$.

All the approaches cited so far consider a  noise model that is independent and identically distributed across tasks, i.e., their MT noise covariance matrices are of the form $\sigma^2\bI$. The approach in \cite{Rakitsch2013} offers a more general solution by introducing a noise covariance matrix which models inter-task noise dependencies. This results in a more realistic model with improved performance compared to the aforementioned alternatives. However, all of these methods have an important drawback in the number of parameters to be inferred. To mitigate this, the models presented in \cite{Bonilla2008, stegle2011efficient, Rakitsch2013} reduce the effective number of parameters of the inter-task covariance matrix by approximating it with a sum of $P$ rank one matrices that are further regularized in the GP model by the noise covariance term.

Additionally, several convolutional models \citep{lee2002, Boyle2005, Alvarez2008, alvarez2011computationally} have emerged, establishing a more sophisticated formulation that is able to model blurred relationships between tasks by the generalization of the MT kernel matrix through a convolution. However, adequate usage requires careful selection of the convolutional kernel in order to make the integral tractable, and the number of parameters must be limited to balance the model's flexibility against its complexity to avoid overfitting issues. Furthermore, this complex design limits the model's interpretability since the inter-task covariance matrices are not explicitly estimated. Efficient versions of these models \cite{alvarez2011computationally} introduce sparse GP formulations able to select $M$ inducing points to reduce the computational cost down to $\mathcal{O}(M^3T)$. 

 \review{ The main challenge for all these approaches (sse also \cite{chen2022multitask, NEURIPS2019_0118a063,nabati2022jgpr}),  lies in the fact that they have to fit a large number of parameters to model all the task relationships and, despite the fact that we can find many ad-hoc implementations that use very accurate optimizers \cite{GPTorch, DEWOLFF2021}, all these approaches are prone to fail in local minima, therefore resulting in suboptimal performance.}

 \review{This work has been inspired by the MTGP probabilistic model featured in \cite{Rakitsch2013}, where the intertask and the noise covariances are explicitly modelled. Our proposal however presents a number of innovations and advantages that are summarized below. 

First, we rely on a new formulation based on a decomposition of the likelihood function into a set of conditional one-output GPs, combined with a hierarchical extension of the conditioned GPs. To our knowledge, this innovation has not been introduced before, and it leads to the following advantages. In the first place, as opposed to previous approaches, the one presented here does not need to use a low rank approximation of the inter-task covariance matrix. This avoids the need to select the corresponding hyperparameter (i.e: approximation rank parameter $P$ in \cite{Rakitsch2013} and others). 

This also means that the present methodology uses a full rank expression of the intertask covariance matrix, with $T(T-1)/2$ parameters. Despite this fact, the number of parameters to be learnt is reduced to $2T$, which is sufficient to recover the full noise and intertask covariance matrices, while previous approaches require $PT$ parameters to recover an intertask covariance of rank $P$.

The complexity of our model is also reduced, which leads to a reduction of the risk of overfitting and, hence, to an improvement in overall predictive performance.

Secondly, this learning approach can be easily adapted to leverage efficient GP libraries, such as Pytorch \cite{GPTorch} or MOGPTK \cite{DEWOLFF2021}, allowing our method to be run on graphical processing units (GPUs) with a computational burden of $\mathcal{O}(TN^2)$. In fact, this implementation is publicly available at \url{https://github.com/OGHinde/Cool_MTGP}. Additionally, in this case, the model also admits an embarrasingly parallel implementation over the tasks to get a computational cost per process of $\mathcal{O}({N^2})$.
}

\section{Introduction to the Multitask Gaussian Processes}
\label{Sect:MT_problem}

Given the set of observations $\bx_i \in \mathbb{R}^D$, $i=1,\cdots,~N$ and a transformation $\bphi(\cdot)$ into a reproducing kernel Hilbert space $\mathcal{H}$ \cite{ShaweTaylor04}, the  general expression for the MT regression problem of estimating $T$ regressors or tasks, $\by_{1:T,i} = \left[ y_{1,i} \cdots y_{T,i}\right]^\top$,  $y_{t,i} \in  \mathbb{R}$ from $\bphi(\bx_i)=\bphi_i$, results in the model
\begin{equation}\label{eq:MT_model}
\by_{1:T,i}   = \bW_{1:T}^\top  \bphi_i + \be_i, 
\end{equation}
$\bW_{1:T} = \left[\bw_{1},\cdots, \bw_{T} \right]$ being a mixing matrix where $\bw_t \in \mathcal{H}$, and $\be_i \in \mathbb{R}^{T}$ representing the model noise. To complete this probabilistic model, the following inter-task noise distribution and weight prior are assumed
\begin{equation}\label{eq:W_1:Tprior}
\begin{split}
p(\be_i)&=\mathcal{N}(\be_i|\bZero,\bSigma_{1:T,1:T})\\
p(\text{vect}(\bW_{1:T}))&= \mathcal{N}(\text{vect}(\bW_{1:T})| \bZero, \bC_{1:T,1:T} \otimes \bSigma_p),
\end{split}
\end{equation}
where $\text{vect}(\cdot)$ is a column-wise vectorization operator and $\otimes$ is the Kronecker product. Matrix $\bSigma_p$ models the covariances between the elements of $\bw_t$ and it is common for all the tasks. This considers that correlation between the noise elements of different tasks is represented through the noise covariance $\bSigma_{1:T,1:T}$, and relationships between  tasks are  modelled with the intertask covariance $\bC_{1:T,1:T}$. 

In order to do Bayesian inference we define the multitask likelihood
\begin{equation}\label{eq:MT_likelihood}
\begin{split}
p\left(\text{vect}({\bY_{1:T}})|  \bPhi,\bW_{1:T}\right) &=\prod_{i=1}^N \mathcal{N}(\by_{1:T,i}|{\bar \by}_{1:T,i},\bSigma_{1:T,1:T}) =\\ &=\mathcal{N}(\text{vect}(\bY_{1:T})|\text{vect}({\bar \bY}_{1:T}),\bSigma_{1:T,1:T}\otimes \bI)
\end{split}
\end{equation}
where $\bPhi=\left[\bphi_{1},\cdots, \bphi_{N} \right] $, $\bY_{1:T}=\left[\by_{1:T,1},\cdots, \by_{1:T,N} \right]$, and the bar notation $\bar{\bu}$ denotes the expectation of any random variable $\bu$; in particular, $\bar{\bY}_{1:T}={\bW}^\top_{1:T}\bPhi$. 
Now, we can obtain the  marginal likelihood (marginalizing the influence of $\bW$):
\begin{equation}\label{eq:MT_marginal_likelihood}
\begin{split}
    &p\left(\text{vect}({\bY_{1:T}})|  \bPhi, \bC_{1:T,1:T}, \bSigma_{1:T,1:T} \right) \\ &=\mathcal{N}(\text{vect}(\bY_{1:T})|\bZero,  \bC_{1:T,1:T}\otimes \bPhi^{\top}\bSigma_p\bPhi +    \bSigma_{1:T,1:T}\otimes \bI)
    \end{split}
\end{equation}

The estimation of matrices  $\bSigma_{1:T,1:T}$ and  $\bC_{1:T,1:T}$ is obtained through the maximization of \eqref{eq:MT_marginal_likelihood}. Finally, the predictive posterior,  $\bbf_{1:T}^*=\left[ f_{1}^* \cdots f_{T}^* \right]^\top$,  for test sample $\bphi^*$ is constructed as:
\begin{equation}\label{eq:MTGP_post_mean_cov}
\begin{split}
& p(\bbf_{1:T}^*|\bphi^*, \by_{1:T}, \bPhi) = \mathcal{N}(\bbf_{1:T}^*|{\bar \bbf}_{1:T}^*, \bC^{*} )\\
&{\bar \bbf}_{1:T}^* = \left(\bC_{1:T,1:T} \otimes \bk_{*}^\top\right)  \left( \bC_{1:T,1:T} \otimes \bK  + \bSigma_{1:T,1:T} \otimes  \bI \right)^{-1}  \text{vect}({\bY_{1:T}})\\
& \bC^{*}= \bC_{1:T,1:T} \otimes k_{**}\\
&-\left(\bC_{1:T,1:T} \otimes \bk_{*}^\top\right)  \left( \bC_{1:T,1:T} \otimes \bK  + \bSigma_{1:T,1:T} \otimes  \bI \right)^{-1} \left(\bC_{1:T,1:T} \otimes \bk_*\right),
\end{split}
\end{equation}
where vector $\bk_*=\bPhi^\top \bSigma_p \bphi^*$  contains the dot products between the test sample $\bphi^*$ and the training dataset $\bPhi$,  $\bK=\bPhi^\top \bSigma_p \bPhi$ is the matrix of dot products between data, and $k_{**}=\bphi^{*\top} \bSigma_p \bphi^*$. 

So far, this formulation extends the model of \cite{Bonilla2008} and \cite{stegle2011efficient} and is formally identical to \cite{Rakitsch2013}, which introduces the noise covariance. The underlying limitation of these works lies in the fact that the number of parameters to be optimized grows with  $T^2$, and therefore a rank-$P$ approximation of the form $\sum_p \lambda_p \bu_p \bu_p^\top + s^2\bI$ is used to model matrices $\bC_{1:T,1:T}$ and $\bSigma_{1:T,1:T}$. This reduces the number of parameters to $2TP$, but imposes the need of selecting a suitable value for parameter $P$.




\section{Parameter learning through conditional one-output likelihood for MTGPs}

Here, we introduce a novel methodology based on a conditional one-output likelihood MTGP (Cool-MTGP) where the previous MTGP formulation is decomposed into a set of $T$ conditioned one-output GPs. 
This methodology reduces the number of parameters to be learnt to twice the number of tasks $T$ without assuming any low rank approximation and the adjustment of additional hyperparameters. 

\subsection{MT likelihood as a product of conditional one-output distributions}

Let us consider a model whose output corresponding to input $\bphi_i$ in the $t$-th  task is given by a linear combination of both the input data and the output of the previous tasks, ${\bf y}_{1:t-1,i}$, i.e.,
\begin{equation}\label{eq:MT_likelihood_chain}
{y}_{t,i}= \bphi ^{\top}_i\bw_{\bx,t} + {\bf y}_{1:t-1,i}^{\top}\bw_{{\bf y},t}  + \epsilon_{t,i}
\end{equation} 
where $\epsilon_{i,t} \sim \mathcal{N}(0,\sigma_t^2)$ is the noise process for task $t$ and the weight of each factorized task is split into two components: weight $\bw_{\bx,t} \in \mathcal{H}$ for the input data and weight $\bw_{{\bf y},t}\in \mathbb{R}^{t-1}$ for the previous tasks. This model is closely related to the original MTGP described in Section \ref{Sect:MT_problem} since, given the values of $\bw_{\bx,t}$  and $\bw_{{\bf y},t}$, one can recursively recover the original weights $\bw_t$ as:
\begin{equation}
    \bw_t = \bw_{\bx,t} + \bW_{1:t-1} \bw_{\by,t}.
\end{equation}
We can now apply the chain rule of probability to the original joint MT likelihood to factorize it into a set of conditional probabilities, each associated to a conditional one-output GP:
\begin{equation}\label{eq:MTGP_factorized}
\begin{split}
&p\left(\text{vect}({\bY_{1:T}})| \bPhi,\bW_{1:T}\right)=p(\by_{T}|\bY_{1:T-1}, \bPhi,\bw_{\bx,T},\bw_{\by,T})\cdot\\
&\cdot p(\by_{T-1}|\bY_{1:T-2}, \bPhi,\bw_{\bx,T-1},\bw_{\by,T-1})\cdots p(\by_{2}|, \by_{1}, \bPhi,\bw_{\bx,2},\bw_{\by,2})\cdot\\
&\cdot p(\by_{1}|, \bPhi,\bw_{\bx,1}),
\end{split}
\end{equation}  
where the likelihood for each of these conditioned GPs is given by:
\begin{equation}\label{eq:likelihood_t}
p({\by}_{t}|{\bY}_{1:t-1},\bPhi , \bw_{\bx,t},\bw_{{\bf y},t} )=\mathcal{N}(\by_{t}|\bPhi ^{\top}\bw_{\bx,t} + {\bY}_{1:t-1}^{\top}\bw_{{\bf y},t} ,\sigma_t^2 \bI).\\
\end{equation}  
And the prior over the input weight components  $\bw_{\bx,t}$ is defined as:
\begin{equation}\label{eq:prior_wx}
p(\bw_{\bx,t})= \mathcal{N}(\bw_{\bx,t} | \bZero, b_t \bSigma_p),
\end{equation} 
where $\bSigma_p$ assumes a common prior for all tasks scaled by a task-dependent factor $b_t$. 

\begin{figure}[t!]
    \centering
    
    \begin{tabular}{cc}

    \includegraphics[scale=0.60]{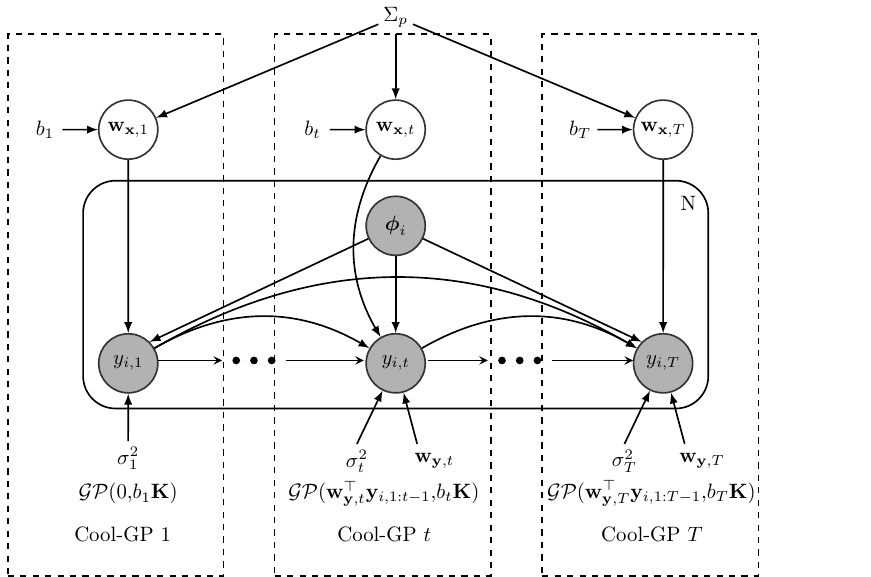} & \hspace{-1cm}
    \includegraphics[scale=0.60]{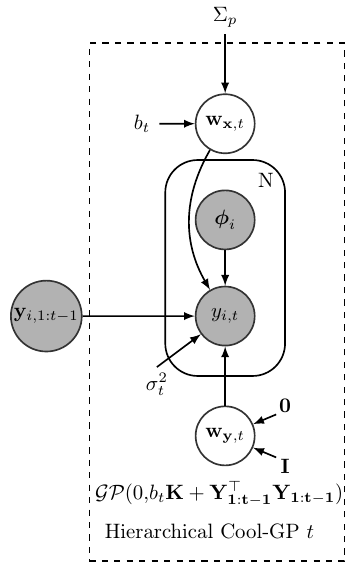} \\
    (a) Global model &\hspace{-1cm} (b) Hierarchical extension
    \end{tabular}
    \caption{Graphical model for the conditional one-output likelihood MTGP model}
    \label{fig:GrapModel}
\end{figure}

This approach assumes that $\bw_{\bx,t}$ are latent variables modelled with a prior distribution, whereas previous task weights $\bw_{{\bf y},t}$ are defined as model parameters (see the graphical model in Figure \ref{fig:GrapModel}(a)); this way, for each task we generate a conditioned one-output likelihood GP (Cool-GP) with mean $\bw_{{\bf y},t}^\top \by_{1:t-1}$ and covariance $b_t\bK$:
 \begin{equation} \label{eq:Cool-GP_t}
    \text{Cool-GP}_t \sim  \mathcal{GP}\left(\bw_{{\bf y},t}^\top \by_{1:t-1}, b_t\bK \right)
  \end{equation}

This guarantees that the model remains Gaussian, allowing us to recover the original MTGP joint likelihood from the set of Cool-GP likelihoods, as defined in \eqref{eq:likelihood_t}. 

\subsection{Parameter learning and model inference}

In order to optimize the model in Figure \ref{fig:GrapModel}(a), we need to learn the input prior factors $\bb_{1:T} = \left[b_1,\ldots,b_T\right]$, the noise covariances $\boldsymbol\sigma^2_{1:T} =\left[ \sigma_1^2, \ldots, \sigma_T^2\right]$, the common kernel parameters $\boldsymbol\theta$ and, additionally, the weights of the previous tasks $\bw_{{\bf y},1}, \ldots, \bw_{{\bf y},T}$. To reduce the number of parameters to be learnt we propose two approaches to infer the values of $\bw_{{\bf y},1}, \ldots, \bw_{{\bf y},T}$ with a closed expression instead of  having to learn their values with gradient descent approaches; in this way, the model complexity is reduced to $2T$ parameters plus the kernel parameters to be learnt. 

\subsubsection{A hierarchical approach for Cool-MTGP learning}
\label{sec:hierarchical}

This first approach proposes to solve the Cool-GP, depicted in Figure \ref{fig:GrapModel}(a), by means of a hierarchical methodology based on a two step learning process.

In the first step, we consider the model introduced in Figure \ref{fig:GrapModel}(b) and define a prior distribution for each $\bw_{{\bf y},t}$ with the form 
\begin{equation}\label{eq:prior_wyt_hier1}
p(\bw_{{\bf y},t})= \mathcal{N}(\bw_{{\bf y},t} | \bZero, \bI).
\end{equation}
that will be used, together with the independent prior assigned to $\bw_{\bx,t}$, 
\begin{equation}\label{eq:prior_wx_hier1}
p(\bw_{\bx,t})= \mathcal{N}(\bw_{\bx,t} | \bZero, b_t \bSigma_p),
\end{equation} 
to infer a joint posterior probability of both sets of parameters.  This joint posterior distribution can be found as an extension of the inference applied to the standard GP \cite{rasmussen2006} as
\begin{equation}\label{eq:posterior_wt}
p\left(\left[ \bw_{\bx,t}, \bw_{{\by},t}\right]|\bPhi,\bY_{1:t}\right)= \mathcal{N}\left(\left[ \bw_{\bx,t}, \bw_{{\bf y},t}\right]| \left[ {\bar\bw_{\bx,t}}, {\bar\bw_{{\bf y},t}}\right], \bA_t^{-1}\right)
\end{equation} 
where the inverse $\bA_t$ of the posterior covariance is
\begin{equation}\label{eq:posterior_w_At}
\bA_t = \sigma_t^{-2} 
\left[\begin{array}{c}
\bPhi\\ {\bY}_{1:t-1}
\end{array}
\right]
\left[\begin{array}{c}
\bPhi\\ {\bY}_{1:t-1}
\end{array}
\right]^{\top} + \left[
    \begin{array}{cc}
    b_t \bSigma_p  & \bZero\\
    \bZero & \bI  
    \end{array}
\right]^{-1} 
\end{equation}
Its mean value provides the MAP estimation of the weight vector $\bw_{\by,t}$, given by
\begin{equation}\label{eq:posterior_wyt_hier1}
{\bar\bw_{\by,t}} = \sigma_t^{-2} \bA_t^{-1} {\bY}_{1:t-1}  \by_t.
\end{equation}
 Alternatively, the solution for the dual vector of $\bw_{\by,t}$ has the same formal expression as those of a standard GP \cite{rasmussen2006}, but using a composed kernel matrix $\bK_{\bx\by,t} =  b_t \bK + {\bY}_{1:t-1}^\top {\bY}_{1:t-1}+ \sigma_t^{2} \bI $:
\begin{equation}\label{eq:alpha_def_hier1}
\balpha_{\by,t} = \bK_{\bx\by,t}^{-1} \by_t,
\end{equation}
Equation \eqref{eq:posterior_wyt_hier1} can be redefined as:

\begin{equation}\label{eq:MAP_alpha_hier1}
{\bar\bw_{\by,t}}={\bY}_{1:t-1}\balpha_{\by,t}.
\end{equation}

Note that this process is equivalent to training a GP with zero mean and covariance matrix  $b_t\bK + {\bY}_{1:t-1}^\top {\bY}_{1:t-1}$ (as depicted in the model in Figure \ref{fig:GrapModel}(b)); this way $\bar\bw_{\by,t}$ has the formal expression of a standard GP \cite{rasmussen2006}, where the input data kernel $\bK$ matrix is rescaled by factor $b_t$ plus a linear kernel matrix for the previous tasks outputs.

In the second step of the hierarchical model, we learn the remaining parameter values ($\bb_{1:T}$,  $\boldsymbol\sigma^2_{1:T}$ and $\boldsymbol\theta$) by going back to the original cool-GP of Equation \eqref{eq:Cool-GP_t} and Figure \ref{fig:GrapModel}(a), where  $\bw_{\by,t}$ is substituted by its MAP estimation. 

Then, we can learn the model parameters ($\bb_{1:T}, \boldsymbol\sigma^2_{1:T}, \boldsymbol\theta $) by maximizing the joint log-likelihood over all the tasks, given by
\begin{equation}\label{eq:log_likelihood_hier}
\begin{split}
& \log p \left( \bY_{1:T}| \bPhi, \bar\bw_{\by,1}, \ldots, \bar\bw_{\by,T}, \bb_{1:T}, \boldsymbol\sigma^2_{1:T}, \boldsymbol\theta \right)\\
&     = \sum_{t=1}^T \log  p\left(\by_t|\bPhi, \bY_{1:t-1}, \bar\bw_{\by,t}, b_t, \sigma^2_t, \boldsymbol\theta \right)
   \end{split}
\end{equation}
To obtain each one of the conditional one-output likelihoods at the right side of Equation \eqref{eq:log_likelihood_hier},  we consider the prior $\mathcal{N}(\bw_{\bx,t}|\bZero,b_t\bSigma_p)$ for parameters $\bw_{\bx,t}$ and marginalize the likelihood with respect to $\bw_{\bx,t}$, resulting in a Gaussian distribution  with the expression
\begin{equation}
    p(\tilde{\by}_t|\bPhi,  b_t, \sigma^2_t, \boldsymbol\theta)=\mathcal{N}\left(\tilde\by_t \right|\bZero,b_t\bPhi^\top\bSigma_p\bPhi+\sigma^2_t\bI)=\mathcal{N}\left(\tilde\by_t \right|\bZero,\bK_{\bx,t}),
\end{equation}
where we make use of the notation $\tilde{\by}_t=\by_t-\bar\bw_{\by,t}^\top\bY_{1:t-1}$ and $\bK_{\bx,t}=b_t \bK+\sigma^2_t \bI$, which takes advantage of the fact that term $\bar\bw_{\by,t}^\top\bY_{1:t-1}$ is considered constant and can then be used as a mean subtracted from the random variable $\by_t$. 

Replacing this expression into  \eqref{eq:log_likelihood_hier}, we obtain the joint log-likelihood over all the tasks 
\begin{equation}\label{eq:log_likelihood_hier1}
\begin{split}
& \log p \left( \bY_{1:T}| \bPhi, \bar\bw_{\by,1}, \ldots, \bar\bw_{\by,T}, \bb_{1:T}, \boldsymbol\sigma^2_{1:T}, \boldsymbol\theta \right)
     = \sum_{t=1}^T \log  p\left(\tilde\by_t|\bPhi , b_t, \sigma^2_t, \boldsymbol\theta \right)=\\
   &=\sum_{t=1}^T -\frac{1}{2}\tilde\by_t^\top \bK_{\bx,t}^{-1} \tilde\by_t -\frac{1}{2}\log |\bK_{\bx,t}| -\frac{N}{2}\log 2\pi
   \end{split}
\end{equation}

The criterion for the inference of the parameters consists of the maximization of this log likelihood through gradient ascent with respect to them. The  derivatives with respect to each one of the common parameters $\btheta$ are 
\begin{equation}\label{eq:grad_theta_joint_loglikelihood_hier1}
\begin{split}
& \frac{\partial}{\partial  \theta_j} \log p \left( \bY_{1:T}| \bPhi, \bar\bw_{\by,1}, \ldots, \bar\bw_{\by,T}, \bb_{1:T}, \boldsymbol\sigma^2_{1:T}, \boldsymbol\theta \right) = \\
  &   = \sum_{t=1}^T \left(-\frac{1}{2}\tilde\by_t^\top \bK_{\bx,t}^{-1}  \frac{\partial \bK_{\bx,t}}{\partial  \theta_j}       \bK_{\bx,t}^{-1}  \tilde\by_t -\frac{1}{2} \text{trace}\left(\bK_{\bx,t}^{-1} \frac{\partial \bK_{\bx,t}}{\partial  \theta_j} \right) \right)
\end{split}
\end{equation}

For the task dependent parameters, $\bb_{1:T}$ and $\bsigma^2_{1:T}$, these derivatives are 
\begin{equation}\label{eq:grad_bt_joint_loglikelihood_t_hier1}
\begin{split}
& \frac{\partial}{\partial  b_t} \log p \left( \bY_{1:T}| \bPhi, \bar\bw_{\by,1}, \ldots, \bar\bw_{\by,T}, \bb_{1:T}, \boldsymbol\sigma^2_{1:T}, \boldsymbol\theta \right) = \\
&     = -\frac{1}{2}\tilde\by_t^\top \bK_{\bx,t}^{-1}  \frac{\partial \bK_{\bx,t}}{\partial  b_t}       \bK_{\bx,t}^{-1}  \tilde\by_t -\frac{1}{2} \text{trace}\left(\bK_{\bx,t}^{-1} \frac{\partial \bK_{\bx,t}}{\partial  b_t} \right) 
\end{split}
\end{equation}

\begin{equation}\label{eq:grad_sigmat_joint_loglikelihood_t_hier1}
\begin{split}
& \frac{\partial}{\partial  \sigma^2_t} \log p \left( \bY_{1:T}| \bPhi, \bar\bw_{\by,1}, \ldots, \bar\bw_{\by,T}, \bb_{1:T}, \boldsymbol\sigma^2_{1:T}, \boldsymbol\theta \right) = \\
&     = -\frac{1}{2}\tilde\by_t^\top \bK_{\bx,t}^{-1}  \frac{\partial \bK_{\bx,t}}{\partial  \sigma^2_t}       \bK_{\bx,t}^{-1}  \tilde\by_t -\frac{1}{2} \text{trace}\left(\bK_{\bx,t}^{-1} \frac{\partial \bK_{\bx,t}}{\partial  \sigma^2_t} \right) 
\end{split}
\end{equation}

Note that the derivatives of the common parameters $\btheta$ are the sum of the partial derivatives for each task, which implies maximizing the joint multitask likelihood; whereas the derivatives of task dependent parameters, $\bb_{1:T}$ and $\boldsymbol\sigma^2_{1:T}$, only depend on their associated factorized likelihoods. The algorithm for the inference of the parameters consists of the estimation of parameters $\balpha_{\by,t}$ and $\bw_{\by,t}$ with equations \eqref{eq:alpha_def_hier1}  and \eqref{eq:MAP_alpha_hier1} and, later, the optimization of the log likelihood in equation \eqref{eq:log_likelihood_hier1} with respect to parameters $\boldsymbol\theta, \bb_{1:T}, \boldsymbol\sigma^2_{1:T}$ through gradient ascent with the use of gradients \eqref{eq:grad_theta_joint_loglikelihood_hier1}, \eqref{eq:grad_bt_joint_loglikelihood_t_hier1} and \eqref{eq:grad_sigmat_joint_loglikelihood_t_hier1}. The process must be repeated until some convergence criterion has been reached.


Finally, the dual parameters corresponding to the MAP estimation of the input related parameter vector can be computed as
\begin{equation}\label{eq:MAP_wx_dual}
    \bar\bw_{\bx,t}=\bPhi\balpha_{t}
\end{equation}
where ${\balpha_{t} }= \bK_{\bx,t}^{-1} \left(\by_t-\bar\bw_{\by,t}^\top\bY_{1:t-1} \right)$.
This process is summarized in Algorithm \ref{al:inference_algorithm_hier}. 

\begin{algorithm} 
\SetAlgoLined
\footnotesize
  \caption{Hierarchical\_Cool\_GPs}\label{al:inference_algorithm_hier}
$\boldsymbol\sigma^2_{1:T}, \bb_{1:T} , \btheta, \balpha_{1:T},  \bw_{\by,1}, \ldots, \bw_{\by,T}$ =  Hierarchical\_Cool\_GPs($\bX, \bY_{1:T}, K$)\\

\KwData{$\bX$ (inputs), $\bY_{1:T}$ (multi-output targets), $K$ (covariance function)}

\tcp{Randomly initialize parameters} $\boldsymbol\sigma^2_{1:T}, \bb_{1:T} , \btheta$

\tcp{Compute kernel matrix  of the input data}
$\bK = K(\bX,\bX;\btheta)$ 

\tcp{Infer parameters of factorized GPs}

\While{Likelihood maximum is not reached}{

$\Delta({\boldsymbol\theta}) = 0$ \\

 \For{$t\leftarrow 1$ \KwTo $T$}{
 
 \tcp{Step 1: Obtain MAP estimation of  $\bw_{\by,t}$}
 
 $\balpha_{\by,t} = \left( b_t \bK + {\bY}_{1:t-1}^\top {\bY}_{1:t-1}+ \sigma_t^{2} \bI \right)^{-1} \by_t$
 
 $\bar\bw_{\by,t}=\bY_{1:t-1}\balpha_{\by,t}$

\tcp{Step 2.1: Update independent parameters}
 
$\sigma^2_t,~b_t$ $\leftarrow$ $\sigma^2_t,~ b_t + \mu \displaystyle \frac{\partial}{\partial  (\sigma^2_t,b_t )} \log p \left( \by_{t}| \bPhi, \bY_{1:t-1}, \bar\bw_{\by,t}, b_{t}, \sigma^2_{t}, \boldsymbol\theta, \right)$

 \tcp{Step 2.2: Update gradient for common parameters}
 $\Delta({\boldsymbol\theta})$ $\leftarrow$ 
 $\Delta({\boldsymbol \theta})$+$\mu \displaystyle \frac{\partial}{\partial  \boldsymbol \theta} \log p \left( \by_{t}| \bPhi, \bY_{1:t-1}, \bar\bw_{\by,t}, b_{t}, \sigma^2_{t}, \boldsymbol\theta \right) $
 }
 
 ${\boldsymbol\theta}\leftarrow {\boldsymbol\theta}+\mu\Delta({\boldsymbol\theta})$
 
  $\bK = K(\bX,\bX;\btheta)$ 
 }
 
 \tcp{With inferred parameters, obtain final values of the dual variables}
 \For{$t\leftarrow 1$ \KwTo $T$}{
 $ {\balpha_{t} }= \left( b_t \bK+\sigma^2_t \bI \right)^{-1} \left(\by_t-\bar\bw_{\by,t}^\top\bY_{1:t-1} \right)$
 }

\KwResult{$\boldsymbol\sigma^2_{1:T}$ (noise variances of factorized GPs) , $\bb_{1:T}$ (prior amplitudes), $\btheta$ (kernel or covariance function parameters) , $\balpha_{1:T}$ (dual variables), $\bar\bw_{\by,1}, \ldots, \bar\bw_{\by,T}$ (MAP estimation previous task weights)}
 
\end{algorithm}

The algorithm developed above needs a matrix inversion for the computation of $\balpha_{\by,t}$ and another inversion of $\bK_{\bx,t}$ for the gradient descent and the computation of $\balpha_{t}$. To reduce this computational cost, in the next section we introduce an approximated inference approach that only uses a single learning step, requiring a single matrix to be inverted.

\subsubsection{An approximate approach for Cool-MTGP learning}
\label{sec:approximated}

To simplify the hierarchical learning approach described above, we can assume a prior for both $\bw_{\bx,t}$ and $\bw_{\by,t}$ to learn the model parameters and infer their variables $\bw_{\bx,t}$ and $\bw_{\by,t}$ with a common model. In fact, considering that  $\bw_{\by,t}$ and $\bw_{\bx,t}$ are given by \eqref{eq:prior_wx} and \eqref{eq:prior_wyt_hier1} the model for each Cool-MTGP would consist of a Gaussian Process with zero mean and covariance $b_t\bK + {\bY}_{1:t-1}^\top {\bY}_{1:t-1}$ (the same GP defined by the first step of the hierarchical model as it is shown in Figure \ref{fig:GrapModel}(b)).

So, with this scheme, the joint marginalized MT likelihood would be approximated by the following set of conditional one-output likelihoods\footnote{Note that each cool likelihood is Gaussian, but their covariance is depending on ${\bY}_{1:t-1}$, so their products do not return the equivalent marginalized MT likelihood Gaussian distribution but an approximation to it.} 
\begin{equation}
\begin{split} \label{eq:marginal_likelihood_approx}
& \sum_{t=1}^T \log  p\left({\by}_{t}|\bPhi , {\bY}_{1:t-1}, b_t,  \sigma^2_{t}, \boldsymbol\theta  \right)=\\
   &=-\sum_{t=1}^T \left(\frac{1}{2}\by_t^\top \bK_{\bx\by,t}^{-1} \by_t +\frac{1}{2}\log |\bK_{\bx\by,t}| +\frac{N}{2}\log 2\pi\right)
   \end{split}
\end{equation}
where $\bK_{\bx\by,t}$ is the full kernel matrix that combines the information of the input data with that of the previous tasks. We can learn the model parameters by maximizing them using the following derivatives with respect to the common parameters $\btheta$: 
\begin{equation}\label{eq:grad_theta_joint_loglikelihood_approx}
\begin{split}
 & \frac{\partial}{\partial  \theta_j} \sum_{t=1}^T \log  p\left({\by}_{t}|\bPhi , {\bY}_{1:t-1}, b_t,  \sigma^2_{t}, \boldsymbol\theta  \right) = \\
    & = \sum_{t=1}^T \left(-\frac{1}{2}\by_t^\top \bK_{\bx\by,t}^{-1}  \frac{\partial \bK_{\bx\by,t}}{\partial  \theta_j}       \bK_{\bx\by,t}^{-1}  \by_t -\frac{1}{2} \text{trace}\left(\bK_{\bx\by,t}^{-1} \frac{\partial \bK_{\bx\by,t}}{\partial  \theta_j} \right) \right) 
\end{split}
\end{equation}
and for the task dependent parameters:
\begin{equation}\label{eq:grad_bt_joint_loglikelihood_t_aapprox}
\begin{split}
&\frac{\partial}{\partial  b_t} \sum_{t=1}^T \log  p\left({\by}_{t}|\bPhi , {\bY}_{1:t-1}, b_t,  \sigma^2_{t}, \boldsymbol\theta  \right) =  \frac{\partial}{\partial  \vartheta_j} \log  p\left({\by}_{t}|\bPhi , {\bY}_{1:t-1}, b_t,  \sigma^2_{t}, \boldsymbol\theta  \right) =\\
   &  = -\frac{1}{2}\by_t^\top \bK_{\bx\by,t}^{-1}  \frac{\partial \bK_{\bx\by,t}}{\partial  b_t}       \bK_{\bx\by,t}^{-1}  \by_t -\frac{1}{2} \text{trace}\left(\bK_{\bx\by,t}^{-1} \frac{\partial \bK_{\bx\by,t}}{\partial  b_t} \right) 
\end{split}
\end{equation}

\begin{equation}\label{eq:grad_sigmat_joint_loglikelihood_t_approx}
\begin{split}
&\frac{\partial}{\partial  \sigma_t^2} \sum_{t=1}^T \log  p\left({\by}_{t}|\bPhi , {\bY}_{1:t-1}, b_t,  \sigma^2_{t}, \boldsymbol\theta  \right) =  \frac{\partial}{\partial  \sigma_t^2} \log  p\left({\by}_{t}|\bPhi , {\bY}_{1:t-1}, b_t,  \sigma^2_{t}, \boldsymbol\theta  \right) =\\
   &  = -\frac{1}{2}\by_t^\top \bK_{\bx\by,t}^{-1}  \frac{\partial \bK_{\bx\by,t}}{\partial  \sigma_t^2}       \bK_{\bx\by,t}^{-1}  \by_t -\frac{1}{2} \text{trace}\left(\bK_{\bx\by,t}^{-1} \frac{\partial \bK_{\bx\by,t}}{\partial  \sigma_t^2} \right) 
\end{split}
\end{equation}

Once the model parameters are learnt, we can use the joint posterior distribution of $\bw_{\bx,t}$ and $\bw_{\by,t}$ (see  \eqref{eq:posterior_wt} and \eqref{eq:posterior_w_At}) to obtain their MAPs estimations as:
\begin{equation}\label{eq:MAP_wx_alpha_approx}
{\bar\bw_{\bx,t}}=\bPhi\balpha_{t}
\end{equation}
\begin{equation}\label{eq:MAP_wy_alpha_approx}
{\bar\bw_{\by,t}}={\bY}_{1:t-1}\balpha_{t}
\end{equation}
where the dual variables $\balpha_{t}$ are
\begin{equation}\label{eq:alpha_def_approx}
\balpha_{t} =\bK_{\bx\by,t}^{-1} \by_t 
\end{equation}

\review{The difference between this approach and the hierarchical approach is how $\bw_{\by,t}$ is treated. In the hierarchical model two inference steps are used, one to estimate the mean of $\bw_{\by,t}$ (see Eq. \eqref{eq:alpha_def_hier1}- \eqref{eq:MAP_alpha_hier1}) and another to obtain the cool-GP; but in the approximate Cool-MTGP, each cool-GP models $\bw_{\by,t}$ as a random variable with a single inference step. The disadvantage of this approach relies in the fact that  the conditional likelihoods are no longer Gaussian and therefore we cannot ensure that the original marginalized MTGP likelihood can be exactly recovered. For this reason, this approach is called \textit{approximate}. }

However, assuming that this criterion is valid, we get several advantages. First, this optimization only requires the inversion of matrix $\bK_{\bx\by,t}$ in both parts of the hierarchic process described in the paper, whereas the exact procedure requires the additional inversion of matrix $\bK_{\bx,t}$ in the first level of the hierarchic process. Besides, if the kernel is linear, the approximate procedure can be implemented with the use of standard univariate GP libraries, and a more sophisticated wrapper that modifies the common parameter inference or a simple cross validation can be used for the common parameters. Both procedures have been compared in the experiments, showing similar results, albeit slightly better for the exact process. Algorithm \ref{al:inference_algorithm_approx} summarizes the main steps of this process.

\begin{algorithm}
\SetAlgoLined
\footnotesize
  \caption{  Approximate\_Cool\_GPs}\label{al:inference_algorithm_approx}
$\boldsymbol\sigma^2_{1:T}, \bb_{1:T} , \btheta, \balpha_{1:T},  \bw_{\by,1}, \ldots, \bw_{\by,T}$ =  Approximate\_Cool\_GPs($\bX, \bY_{1:T}, K$)\\
\KwData{$\bX$ (inputs), $\bY_{1:T}$ (multi-output targets), $K$ (covariance function)}

\tcp{Randomly initialize hyperparameters} $\boldsymbol\sigma^2_{1:T}, \bb_{1:T} , \btheta$

\tcp{Compute kernel matrix  of the input data:}
$\bK = K(\bX,\bX;\btheta)$ 

\tcp{Infer parameters of factorized GPs}

\While{Likelihood maximum is not reached}{

$\Delta({\boldsymbol\theta}) = 0$ \\

 \For{$t\leftarrow 1$ \KwTo $T$}{
 
\tcp{Update independent parameters}
 
$\sigma^2_t,~b_t$ $\leftarrow$ $\sigma^2_t,~ b_t + \mu \displaystyle \frac{\partial}{\partial  (\sigma^2_t,b_t )} \log p \left( \by_{t}| \bPhi, b_{t}, \sigma^2_{t}, \boldsymbol\theta \right)$

 \tcp{Update gradient for common parameters}
 $\Delta({\boldsymbol\theta})$ $\leftarrow$ 
 $\Delta({\boldsymbol \theta})$+$\mu \displaystyle \frac{\partial}{\partial  \boldsymbol \theta} \log p \left( \by_{t}| \bPhi, b_{t}, \sigma^2_{t}, \boldsymbol\theta \right) $
 }
 
 ${\boldsymbol\theta}\leftarrow {\boldsymbol\theta}+\mu\Delta({\boldsymbol\theta})$
 
  $\bK = K(\bX,\bX;\btheta)$ 
 }
 
 \tcp{With inferred parameters, obtain final values of the dual variables and MAP estimation of previous task weights}
 \For{$t\leftarrow 1$ \KwTo $T$}{
 $\balpha_t = \bK_{\bx\by,t}^{-1} \by_t$
 
 $\bar\bw_{\by,t}=\bY_{1:t-1}\balpha_t$
 }
 
\KwResult{$\boldsymbol\sigma^2_{1:T}$ (noise variances of factorized GPs) , $\bb_{1:T}$ (prior amplitudes), $\btheta$ (kernel or covariance function parameters) , $\balpha_T$ (dual variables), $\bar\bw_{\by,1}, \ldots, \bar\bw_{\by,T}$ (MAP estimation previous task weights)}
 
\end{algorithm}

\subsection{Recovering the multitask model}

Despite the fact that intertask and noise covariance matrices $\bC_{1:T,1:T}$ and $\bSigma_{1:T,1:T}$ are not explicit in the conditioned model, they can be recovered from parameters $\bb_{1:T}$,  $\boldsymbol\sigma^2_{1:T}$ and the MAP estimation of the  weights, $\bar\bw_{{\bf y},1}, \ldots, \bar\bw_{{\bf y},T}$ and  $\bW_{\bx,1:T} = \left[\bar\bw_{{\bf x},1}, \ldots, \bar\bw_{{\bf x},T}\right]$.  
The general MTGP formulation considers that the joint MT likelihood, which contains the  noise matrix $\bSigma_{1:T,1:T}$, is given by
\begin{equation}
p(\text{vect}(\bY_{1:T})| \bPhi,\bW_{1:T}) = \mathcal{N}(\by_{1:T}|{\bar \by}_{1:T},\bSigma_{1:T,1:T}\otimes \bI).
\end{equation}
Considering the factorization given by \eqref{eq:MTGP_factorized}, the factorized likelihoods of \eqref{eq:likelihood_t}, and applying the properties of the products of conditional Gaussians (see, e.g. \cite{bishop2006pattern}), we can recursively recover the mean  ${\bar \by}_{1:T}$ and the covariance terms, which leads us directly to the MT noise covariance $\Sigma_{1:T,1:T}$, as:
\begin{equation}
\begin{split}
\label{eq:MT_recovered_likelihood_mean}
{\bar \by}_{t} &= \bPhi ^{\top}{\bar\bw_{t}} = \bPhi ^{\top}{\bw_{\bx,t}} + {\bar \by}_{1:t-1}^{\top} \\
\bSigma_{t,t} &= \sigma_t^2 + \bar\bw_{{\bf y},t}^{\top} \bSigma_{1:t-1,1:t-1}  \bar\bw_{{\bf y},t}\\ 
\bSigma_{t,1:t} & = \bSigma_{1:t,t}^\top = \bar\bw_{{\bf y},t}^{\top} \bSigma_{1:t-1,1:t-1}\\
\end{split}
\end{equation}
where it is assumed that  $\bSigma_{1,1} = \sigma_1^2$  and ${\bar \by}_{1} = \bPhi ^{\top}{\bw_{\bx,1}}$.

For the purpose of recovering covariance matrix $\bC_{1:T,1:T}$, two elements are needed:
(1) An expression relating the multitask weights $\bW_{1:T}$ to the input related weights $\bw_{\bx,t}$; and, (2) the joint prior of all $\bw_{\bx,t}$ parameters. We assume here that the values of $\bb_{1:T}$ and the MAP value of the previous task weights $\bar\bw_{\by,1}, \ldots, \bar\bw_{\by,T}$ have been already learnt using any of the MTGP approaches proposed above.

To obtain an expression of $\bW_{1:T}$ as a function of the input parameters $\bw_{\bx,t}$, we use
\begin{equation}
{\bar \by}_{t} = \bPhi ^{\top}{\bar\bw_{t}} = \bPhi ^{\top}{\bw_{\bx,t}} + {\bar \by}_{1:t-1}^{\top}{\bar{\bw}_{{\bf y},t}}
\end{equation}
to get:
\begin{equation}
    \bw_t = \bw_{\bx,t} + \bW_{1:t-1}^\top \bar\bw_{\by,t},
\end{equation}
that extended for $t=1, \ldots, T$, leads to the following equation system
\begin{equation}\label{eq:weight_vector_MT_model_apend}
    \bW_{1:T}^\top=\left(\bI-\bar\bW_\by \right)^{-1}\bW_{\bx,1:T}^\top
\end{equation}
where $\bW_{\bx,1:T} =  \left[\bw_{\bx,1},\cdots, \bw_{\bx,T} \right]$ and 
\begin{equation} \label{eq:wy_posterior_matrix_apend}
    \bar\bW_{\by}=\left[
    \begin{array}{ccccc}
         0 & 0 & \cdots & 0 & 0\\
         \bar{\bw}_{\by,2}[1] & 0 & \cdots & 0 &0\\
         \vdots & \vdots & \ddots & \vdots  & \vdots\\
         \bar{\bw}_{\by,T-1}[1] & \bar{\bw}_{\by,T-1}[2] & \cdots & 0 & 0\\
         \bar{\bw}_{\by,T}[1] & \bar{\bw}_{\by,T}[3] & \cdots & \bar{\bw}_{\by,T}[T-1] & 0
    \end{array}
    \right],
\end{equation}
where $\bar\bw_{\by,t}[k]$ is component $k$ of vector $\bar\bw_{\by,t}$.

In order to identify the joint prior of all $\bw_{\bx,t}$ parameters, we formulate it as the following distribution
\begin{equation}\label{eq:prior_Wx_appendix}
p\left(\text{vect}(\bW_{\bx,1:T})\right) =    \mathcal{N} \left(\text{vect}(\bW_{\bx,1:T})|\bZero,\bB_{1:T,1:T} \otimes\bSigma_p \right)
\end{equation}
with matrix $\bB_{1:T,1:T}$ being a cross correlation between tasks. Taking into account that the diagonal terms of $\bB_{1:T,1:T}$  are $b_{tt}=b_t$ (they have been already learnt -- see Sections \ref{sec:hierarchical} and \ref{sec:approximated}), the remaining cross terms can be expressed  through  the correlation coefficient 
\begin{equation}\label{eq:b_tt_apend}
    b_{tt'}=\sqrt{b_t b_{t'}}\hat\rho_{t,t'}
\end{equation}
where $\hat\rho_{t,t'}$ can be estimated as:
\begin{equation}\label{eq:rho_tt_apend}
\hat\rho_{t,t'} =  \frac{\bw_{\bx,t}^\top \bw_{\bx,t'}}{ \vert\vert \bw_{\bx,t} \vert\vert \vert\vert \bw_{\bx,t'}\vert\vert}
\end{equation}

Since this computation depends on variables $\bw_{\bx,t}$ and $\bw_{\bx,t'}$, we can carry it out by either generating samples from their posterior distribution and obtaining the values of $\bB_{1:T,1:T}$ with a Monte Carlo approximation, or by directly considering that $\bW_{\bx,1:T}$ are given by their MAP estimations. In fact, if we consider the latter approach, the calculation of the terms of $\bB_{1:T,1:T}$ can be expressed in a more compact form as:
\begin{equation}\label{eq:b_tt_red}
    b_{tt'}=\sqrt{b_t b_{t'}}\hat\rho_{t,t'} 
    = \frac{\sqrt{b_t b_{t'}} \balpha_t^\top \bK \balpha_{t'}}{ \sqrt{ \balpha_t^\top \bK \balpha_{t} \balpha_{t'}^\top \bK \balpha_{t'}}}.
\end{equation}
where $\balpha_t=\bK_{\bx,t}^{-1} \left(\by_t-\bar\bw_{\by,t}^\top\bY_{1:t-1} \right)$ with $\bar\bw_{\by,t}$ given by \eqref{eq:posterior_wyt_hier1} in the hierarchical Cool-MTGP method, or $\balpha_t=\bK_{\bx\by,t}^{-1} \by_t$ in the approximate approach.

Finally, the knowledge of matrix $\bB_{1:T,1:T}$ together with equation  \eqref{eq:weight_vector_MT_model_apend} leads to the expression of the multitask covariance matrix 
\begin{equation}
\begin{split}
    \bC_{1:T,1:T} \otimes \bSigma_p &= \mathbb{E}\left\lbrace \text{vect}(\bW_{1:T})\text{vect}(\bW_{1:T})^\top  \right\rbrace\\
    &= \mathbb{E}\left\lbrace \text{vect}\left( (\bI-\bW_\by )^{-1}\bW_{\bx,1:T}^\top\right) \text{vect}\left((\bI-\bW_\by )^{-1}\bW_{\bx,1:T}^\top\right)^\top \right\rbrace\\
    &= \left(\bI-\bW_\by \right)^{-1} \mathbb{E}\left\lbrace \text{vect}(\bW_{\bx,1:T}^\top) \text{vect}(\bW_{\bx,1:T}) \right\rbrace \left(\left(\bI-\bW_\by \right)^{-1}\right)^\top\\
\end{split}
\end{equation}
where, by equation \eqref{eq:prior_Wx_appendix}, we see that 
\begin{equation}\label{eq:CTT_kron_Sigmap}
    \bC_{1:T,1:T} \otimes \bSigma_p 
    = \left(\bI-\bW_\by \right)^{-1} \left( \bB_{1:T,1:T} \otimes \bSigma_p  \right)  \left(\left(\bI-\bW_\by \right)^{-1}\right)^\top
\end{equation}
and, finally, using the Kronecker product properties, it reduces to
\begin{equation} \label{eq:C_tt_apend}
    \bC_{1:T,1:T} = \left(\bI-\bar\bW_\by \right)^{-1} \bB_{1:T,1:T}   \left(\left(\bI-\bar\bW_\by \right)^{-1}\right)^\top\\
\end{equation}
The summary of the process is in Algorithm \ref{al:recover_MTGP}.

\begin{algorithm}
\SetAlgoLined
\footnotesize
\caption{Cool\_MTGP\_Learning}\label{al:recover_MTGP} 
$\bSigma_{1:T,1:T}, \bC_{1:T,1:T}, \btheta$ = Cool\_MTGP\_Learning ($\bX, \bY_{1:T}, K$, Type)\\

\KwData{$\bX$ (inputs), $\bY_{1:T}$ (multi-output targets), $K$ (covariance function), Type (indicator for hierarchical or approximate Cool-GPs approach)}

\tcp{Train a model of Cool-GPs}
\If {Type == hierarchical}{
$\boldsymbol\sigma^2_{1:T}, \bb_{1:T} , \btheta, \balpha_{1:T}, \bar\bw_{\by,1}, \ldots, \bar\bw_{\by,T}$ = Hierarchical\_Cool\_GPs($\bX, \bY_{1:T}, K$)
}
\If {Type== approximate}{
$\boldsymbol\sigma^2_{1:T}, \bb_{1:T} , \btheta, \balpha_{1:T}, \bar\bw_{\by,1}, \ldots, \bar\bw_{\by,T}$ =  Approximate\_Cool\_GPs($\bX, \bY_{1:T}, K$)
}

 
  \tcp{Compute multitask covariance of noise}
 $\Sigma_{1,1} = \sigma_1^2$\\
 \For{$t\leftarrow 2$ \KwTo $T$}{
 $\bSigma_{t,t} = \sigma_t^2 + \bar\bw_{{\bf y},t}^{\top} \bSigma_{1:t-1,1:t-1}  \bar\bw_{{\bf y},t} $ \\
$ \bSigma_{t,1:t} = \bSigma_{1:t,t}^\top = \bar\bw_{{\bf y},t}^{\top} \bSigma_{1:t-1,1:t-1}$\\
}
 
 \tcp{Compute multitask weights of the prior}

$\bC_{1:T,1:T} = \left(\bI-\bar\bW_\by \right)^{-1} \bB_{1:T,1:T}   \left(\left(\bI-\bar\bW_\by \right)^{-1}\right)^\top$

where  $\bar\bW_\by$ is given by \eqref{eq:wy_posterior_matrix_apend}, $\bK = K(\bX,\bX;\btheta)$ and entries $t,t'$ of $\bB_{1:T,1:T}$ are given by 

$ b_{tt'}=\displaystyle \frac{\sqrt{b_t b_{t'}}\balpha_t^\top \bK \balpha_{t'}}{ \sqrt{ \balpha_t^\top \bK \balpha_{t} \balpha_{t'}^\top \bK \balpha_{t'}}}$,

\KwResult{$\bSigma_{1:T,1:T}$ (noise covariance matrix) and $\bC_{1:T,1:T}$ (intertask covariance matrix)}

\end{algorithm}

One might be concerned that this reconstruction process, and therefore the overall performance of the model depends on the order in which the tasks are assigned to the Cool-GPs. However, experimental results 
show that the reconstruction of the matrices is consistent for any random permutation of the tasks, confirming that task order shows  no impact.




\section{Predictive multitask model}

Once the noise $\bSigma_{1:T,1:T}$ and intertask $\bC_{1:T,1:T}$ covariance matrices have been obtained using any of the described methods, we are ready to apply the general predictive model of the multitask Gaussian process from equations \eqref{eq:MTGP_post_mean_cov} (see Section \ref{Sect:MT_problem}) to obtain the predictive model as describes the Algorithm \ref{al:PredictiveMTGP_algorithm}.

\begin{algorithm}
\footnotesize
\SetAlgoLined
\caption{Cool\_MTGP\_Predictive}
\label{al:PredictiveMTGP_algorithm}

${\bar \bbf}_{1:T}^*$, $\text{cov}(\bbf_{1:T}^*)$ = Cool\_MTGP\_Predictive ($\bX, \bY_{1:T}, K, Type, \bx^*$)
\\

\KwData{$\bX$ (inputs), $\bY_{1:T}$ (multi-output targets), $K$ (covariance function), Type (indicator for hierarchical or approximate Cool-GPs approach), $\bx^*$ (test input)}

\tcp{Inference over MTGP model}

$\bSigma_{1:T,1:T}, \bC_{1:T,1:T}, \btheta$ = 
Cool\_MTGP\_Learning ($\bX, \bY_{1:T}, K$,Type )

\tcp{Compute kernel matrices}
$\bK = K(\bX,\bX;\btheta)$ 

$\bk_* = K(\bX,\bx^*;\btheta)$ 

$\bk_{**} = K(\bx^*,\bx^*;\btheta)$ 

\tcp{Predictive Mean and Covariance}
$ {\bar \bbf}_{1:T}^* = \left(\bC_{1:T,1:T} \otimes \bk_{*}^\top\right)  \left( \bC_{1:T,1:T} \otimes \bK)  + \bSigma_{1:T,1:T}^{-1} \otimes  \bI \right)^{-1}  \text{vect}({\bY_{1:T}}) $

$ \text{cov}(\bbf_{1:T}^*) = \left(\bC_{1:T,1:T} \otimes \bk_{**} \right) - \left(\bC_{1:T,1:T} \otimes \bk_{*}^\top\right)  \left( \bC_{1:T,1:T} \otimes \bK  + \bSigma_{1:T,1:T}^{-1} \otimes  \bI \right)^{-1} \left(\bC_{1:T,1:T} \otimes \bk_*\right) $

\KwResult{ ${\bar \bbf}_{1:T}^*$  (predictive mean), $\text{cov}(\bbf_{1:T}^*)$ (predictive covariance)}

\end{algorithm}


\section{Experimental Results}

\subsection{Synthetic benchmark}
We have carried out a synthetic data experiment to compare the performance of all the considered MTGP algorithms against a ground truth model. Following the general model of Section \ref{Sect:MT_problem}, a data set has been drawn from likelihood function 
$ p(\by | \bX) = \mathcal{N}(\by|\bZero,\bC_{1:T,1:T}\otimes\bK + \bSigma_{1:T,1:T}\otimes\bI),$ where $\bK = \bX^\top \bX$ and the intertask and noise covariance matrices follow the low rank form $\bC_{1:T,1:T} = \sum_{r=1}^R \bc_r  \bc_r^\top$ (and similarly for $\bSigma_{1:T,1:T}$). We have created datasets for three scenarios in which the $\bC_{1:T,1:T}$ and $\Sigma_{1:T,1:T}$ matrices were generated with $R$ values of $=5$, $10$ and $15$. In all cases, $T=15$ tasks, $N=200$ samples and $10$ iterations were run with randomly split training and test partitions with 100 samples each.

We compare both the hierarchical (HCool-MT) and approximate ($\sim$Cool-MT) versions of the proposed model to the standard MTGP (Std-MT) of \cite{Bonilla2008} and the extension introduced in \cite{Rakitsch2013}, which includes a noise matrix ($\Sigma$-MT). Since these methods require the selection of parameter $P$, we have analyzed three values: one equal to $R$ (the ideal case), a value of $P$ smaller than $R$ and, where possible, a value of $P$ greater than $R$. Additionally, a ground-truth model that uses the true intertask and noise covariance matrices is included, as well as a set of $T$ independent GPs. Predictive performance is measured with the mean square error (MSE) averaged over all the tasks.


\begin{figure}[t!]
    \centering
    \footnotesize
    \begin{tabular}{ccc}
      \hspace{-0.2cm}  \includegraphics[scale=0.36]{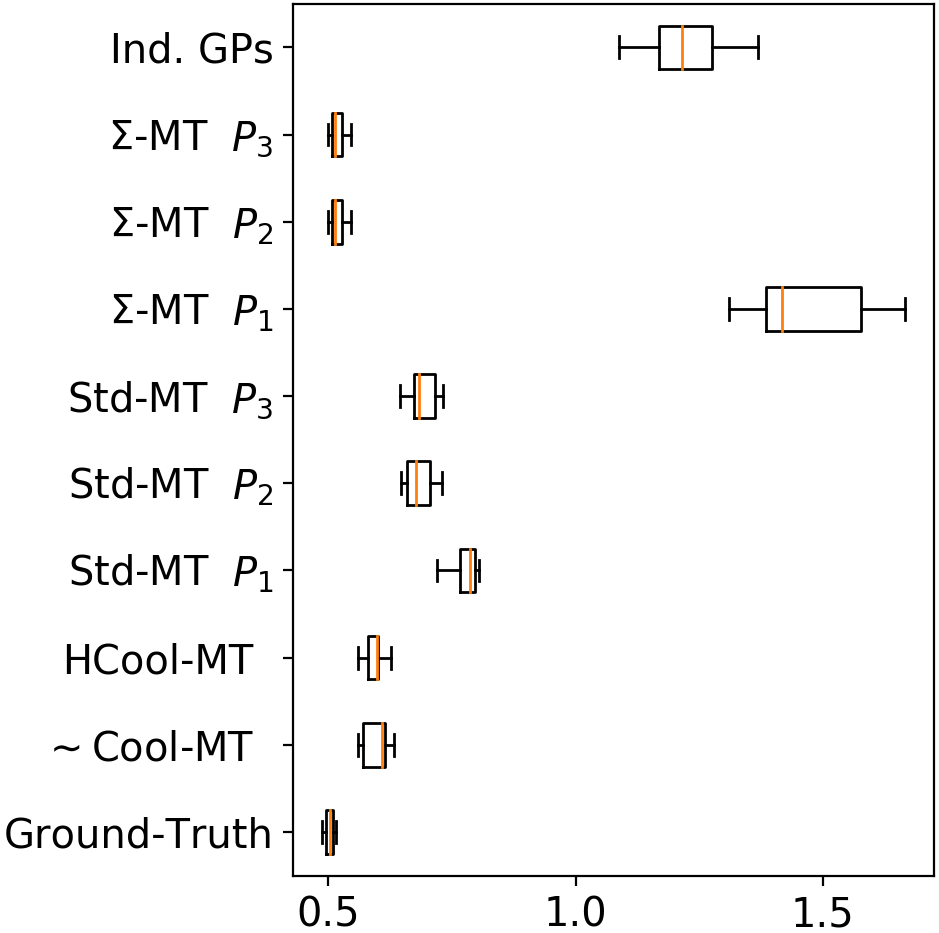}&\hspace{-0.15cm}
        \includegraphics[scale=0.36]{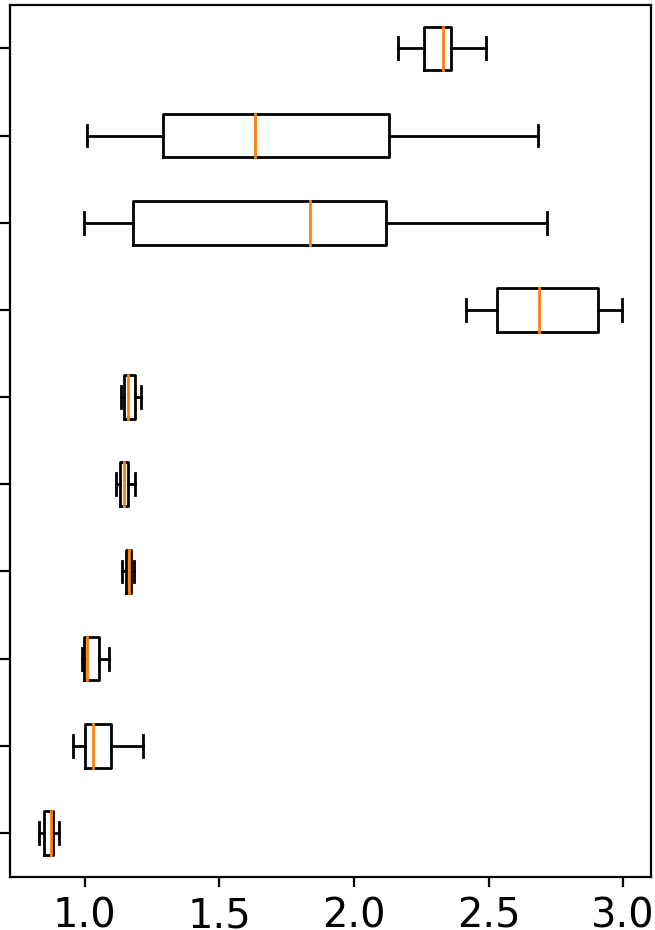}&\hspace{-0.25cm}
        \includegraphics[scale=0.36]{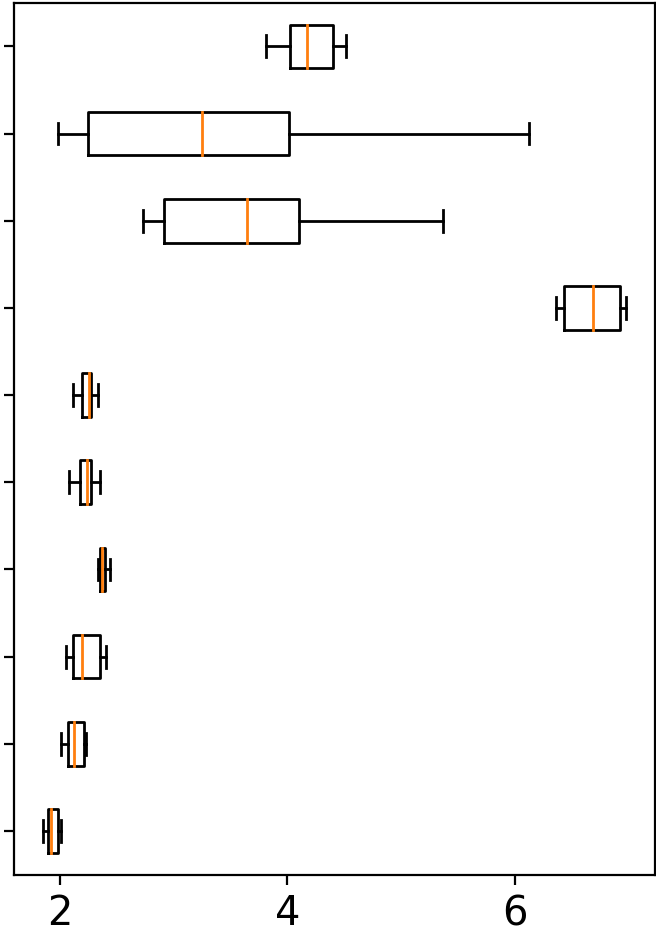}\\
     \hspace{1cm}(a) $R=5$ & (b) $R=10$  &(c) $R=15$\\
     \hspace{1cm} $P_1=3$, $P_2=5$, $P_3=10$&  $P_1=5$, $P_2=10$, $P_3=15$  &  $P_1=5$, $P_2=10$, $P_3=15$
    \end{tabular}
    \caption{MSE for all models of the synthetic experiment and different values of the matrix rank $R$ of the generative model and parameter $P$ for Std-MT and $\Sigma$-MT.}
    \label{fig:synth-boxplot}
\end{figure}

\def\Scale2{0.18}
\begin{figure}[t!]
\centering
\footnotesize
\begin{tabular}{ccccccc}
        \hspace{-0.1cm}\rotatebox{90}{\hspace{0.6cm}$\bC_{1:T,1:T}$} &
        \hspace{-0.4cm}\includegraphics[scale=\Scale2]{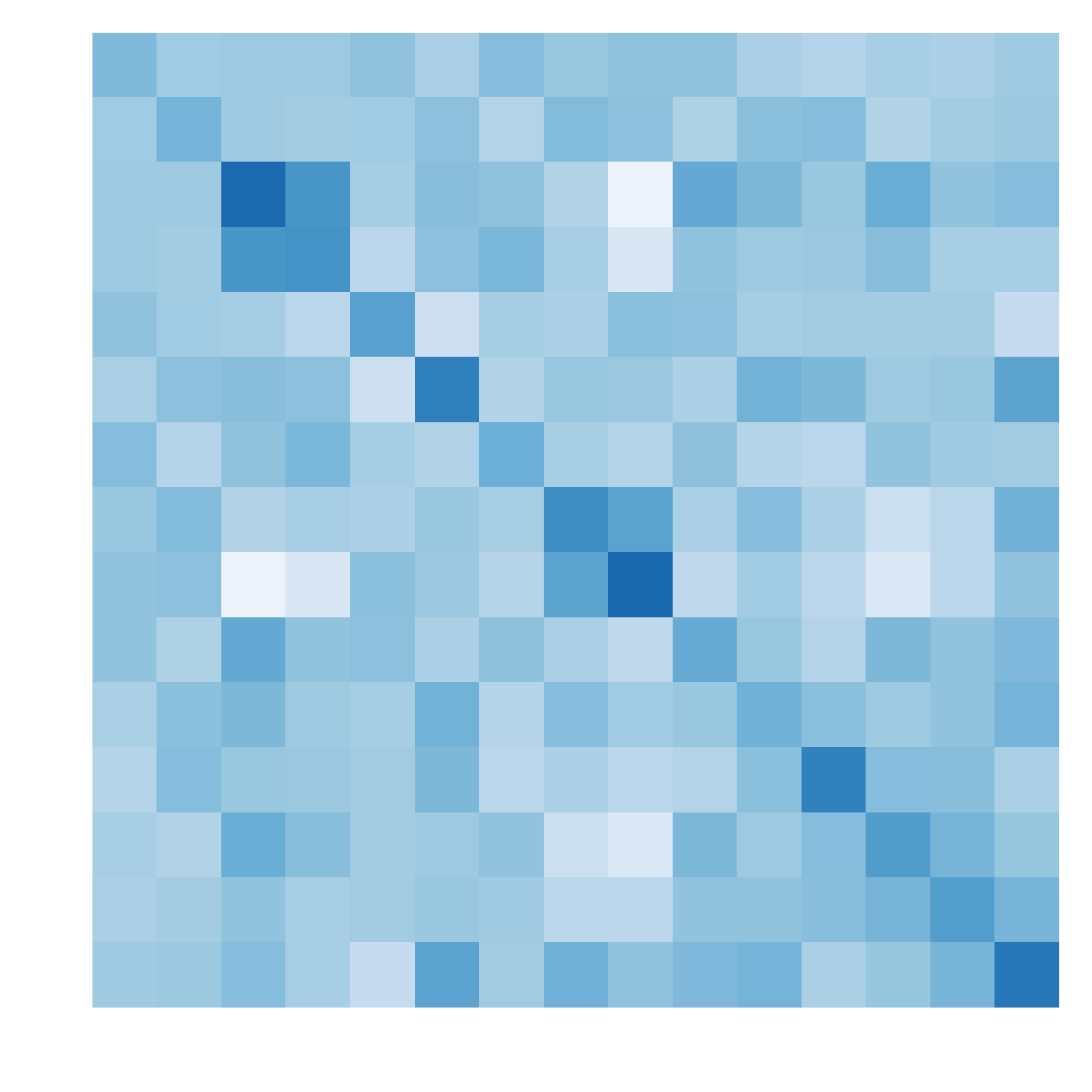}&
        \hspace{-0.4cm}\includegraphics[scale=\Scale2]{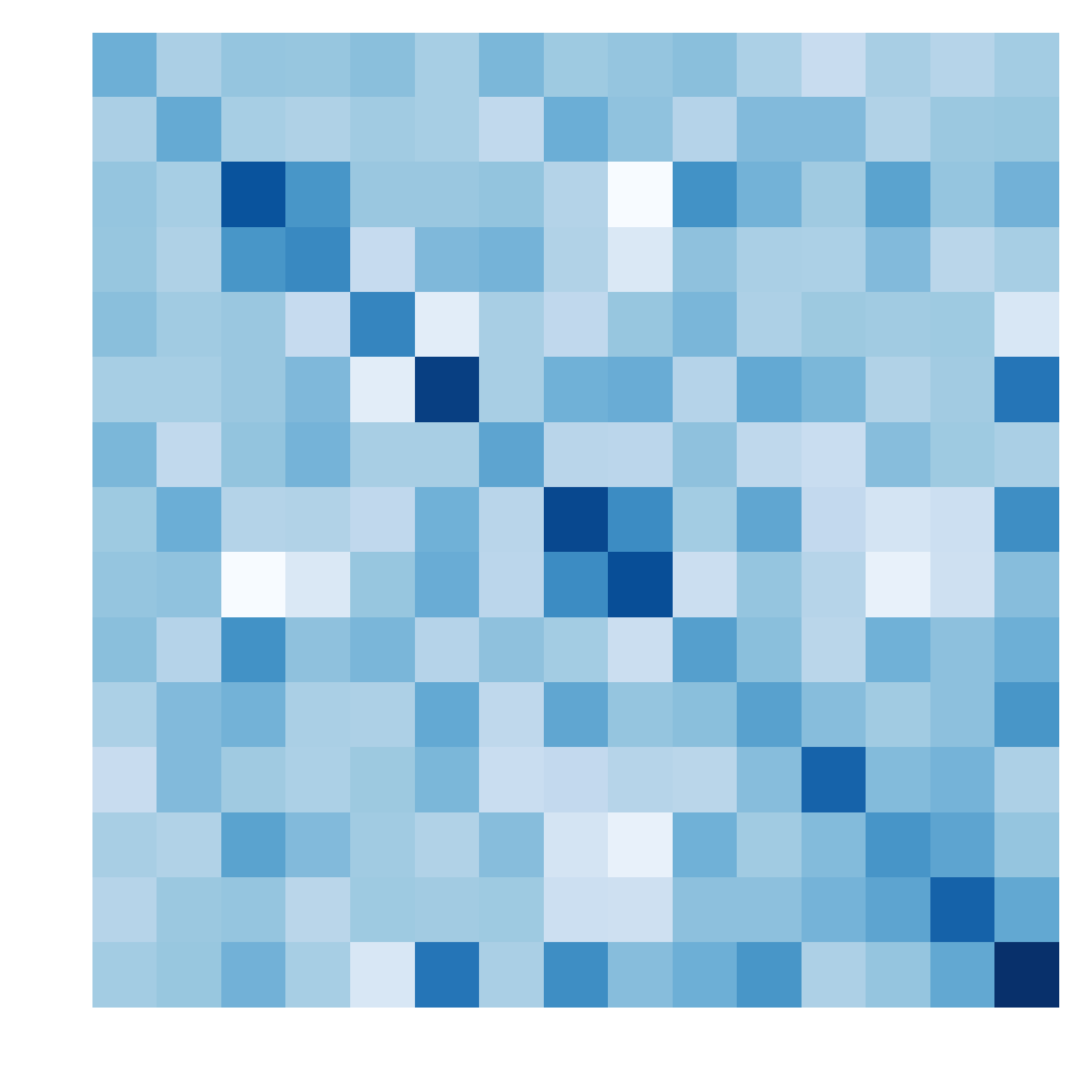}&
        \hspace{-0.4cm}\includegraphics[scale=\Scale2]{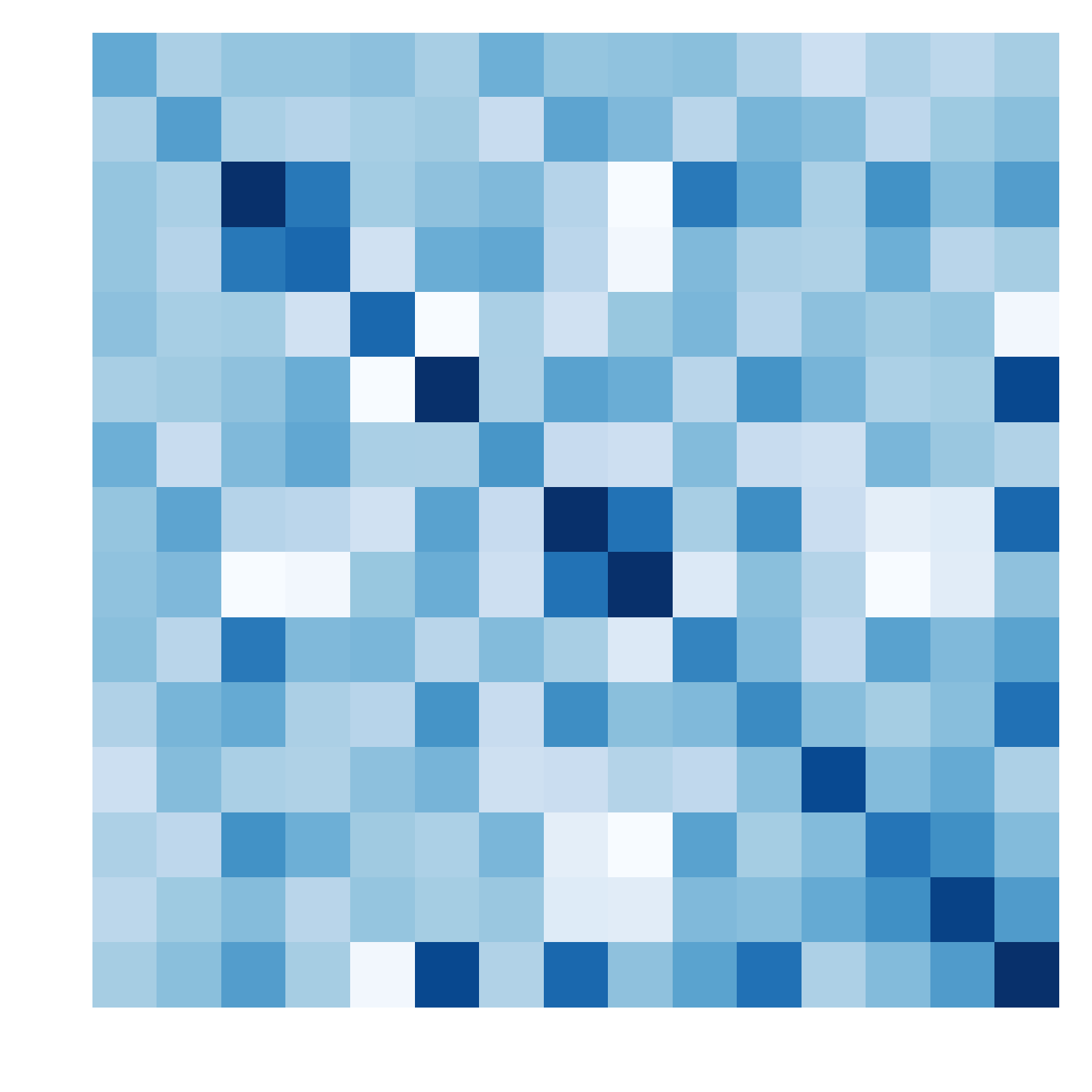}&
        \hspace{-0.4cm}\includegraphics[scale=\Scale2]{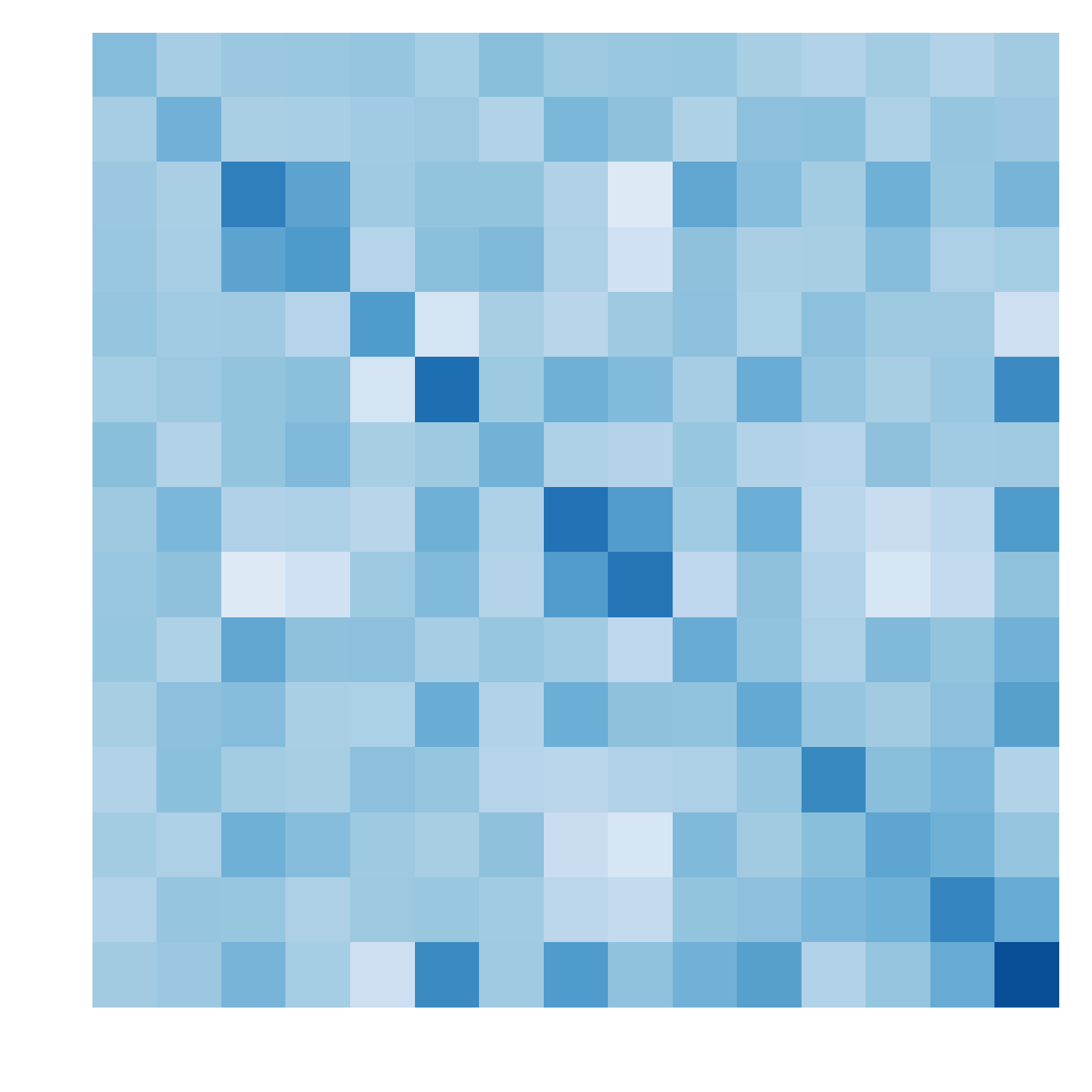}&
        \hspace{-0.4cm}\includegraphics[scale=\Scale2]{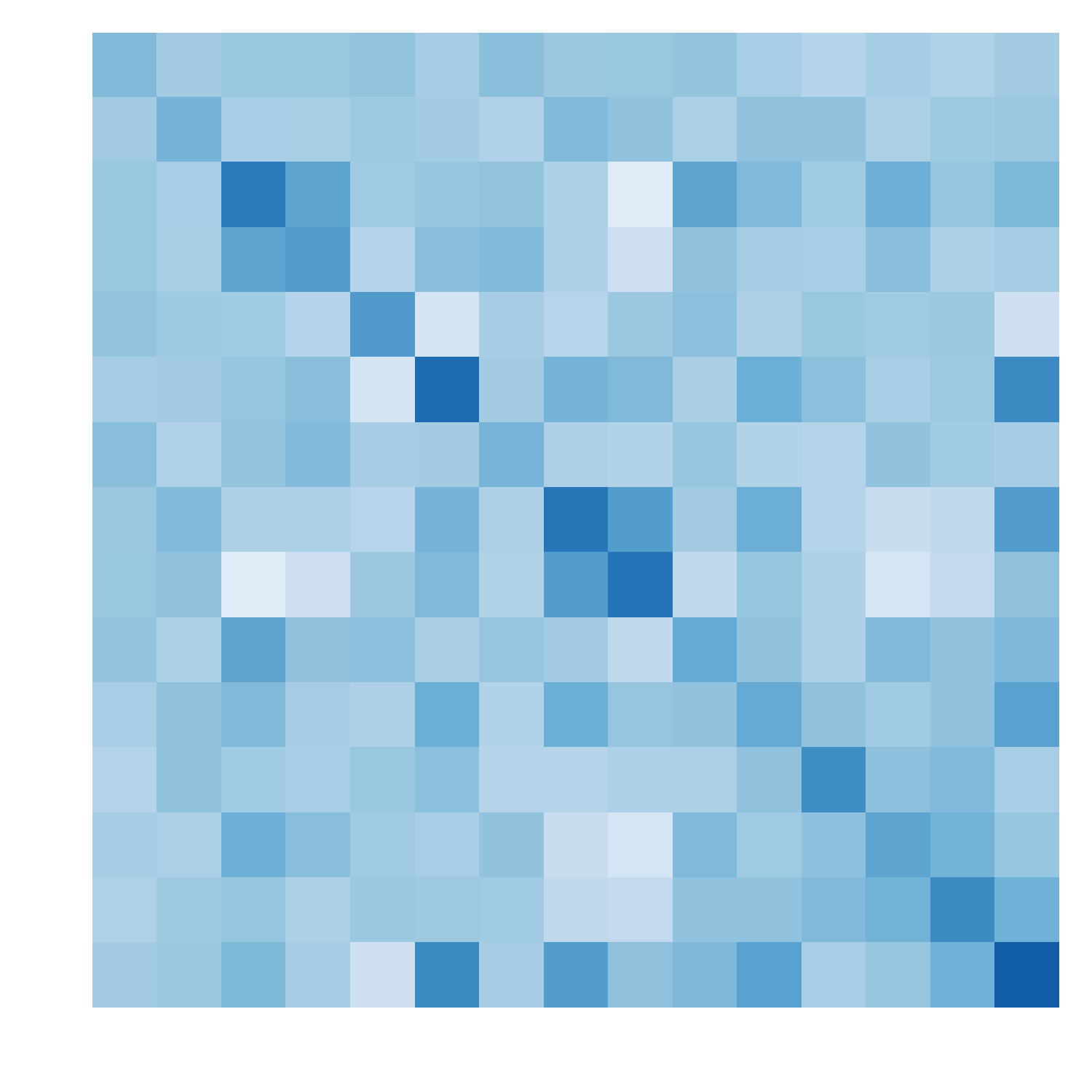}&
        \hspace{-0.3cm}\includegraphics[scale=\Scale2]{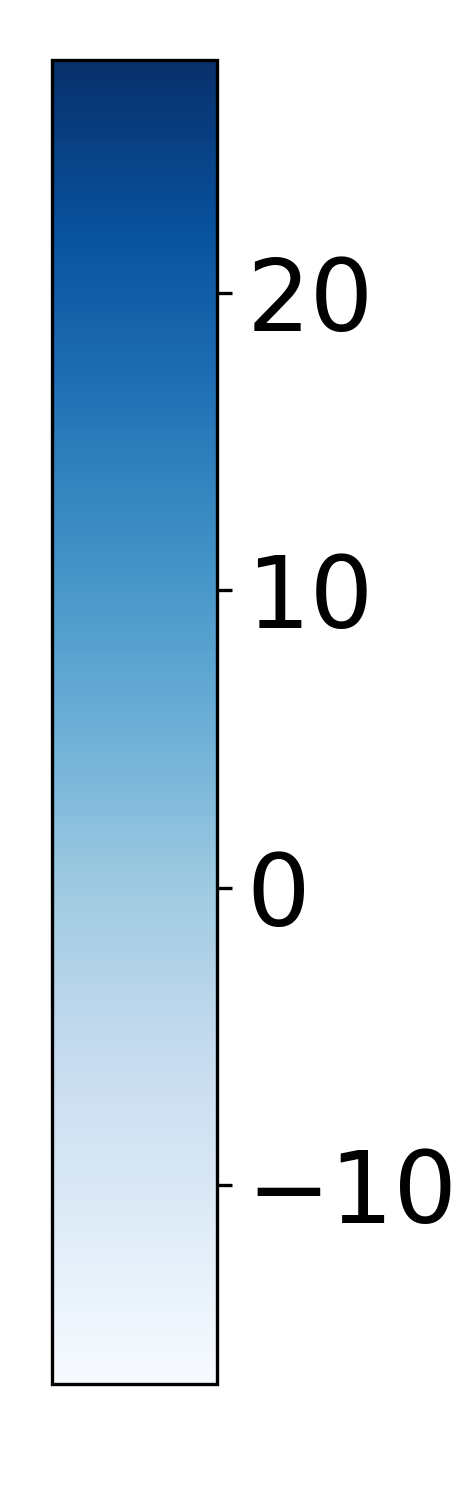} \\
        \hspace{-0.1cm}\rotatebox{90}{\hspace{0.6cm}$\bSigma_{1:T.1:T}$}& \hspace{-0.4cm}\includegraphics[scale=\Scale2]{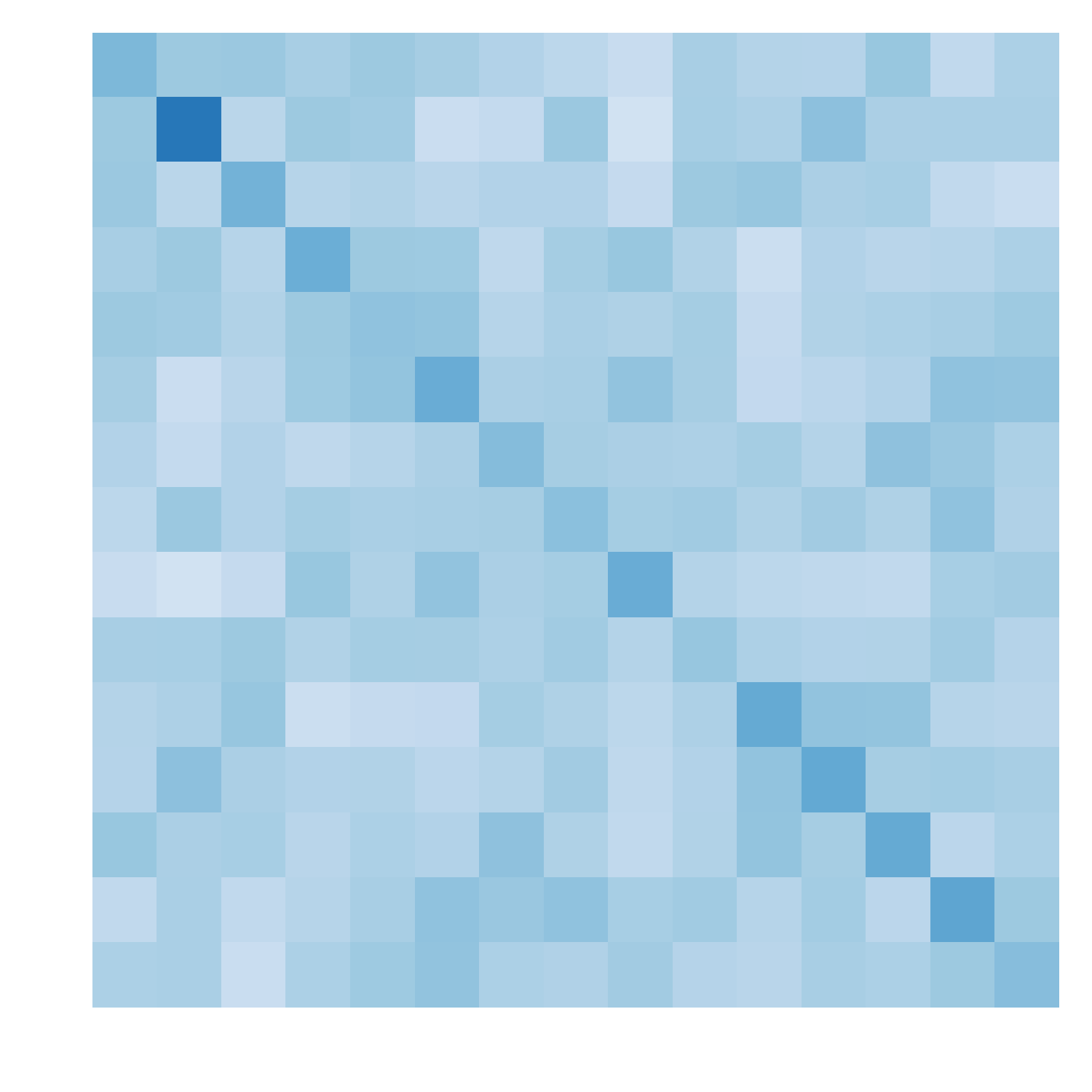}&
        \hspace{-0.4cm}\includegraphics[scale=\Scale2]{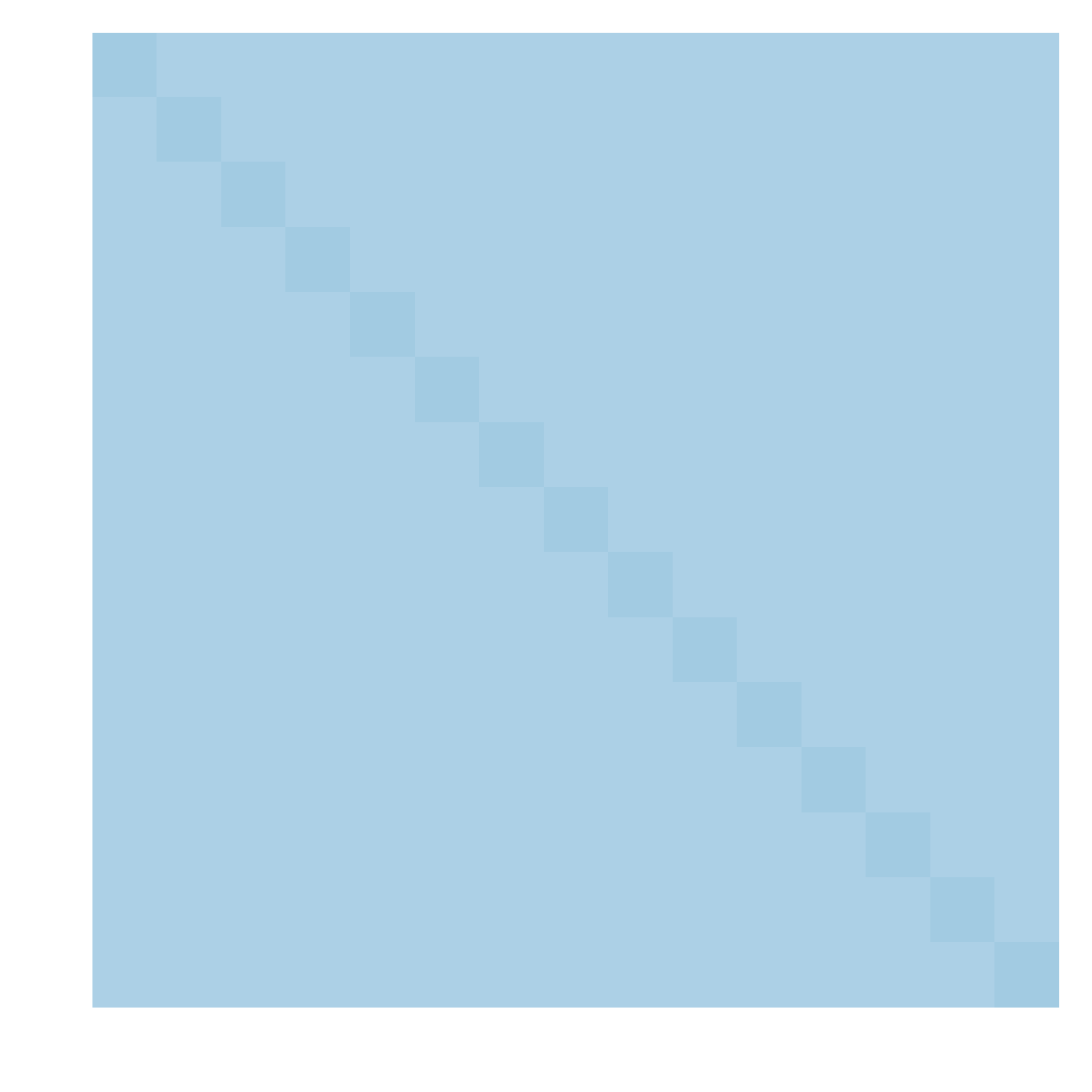}&
        \hspace{-0.4cm}\includegraphics[scale=\Scale2]{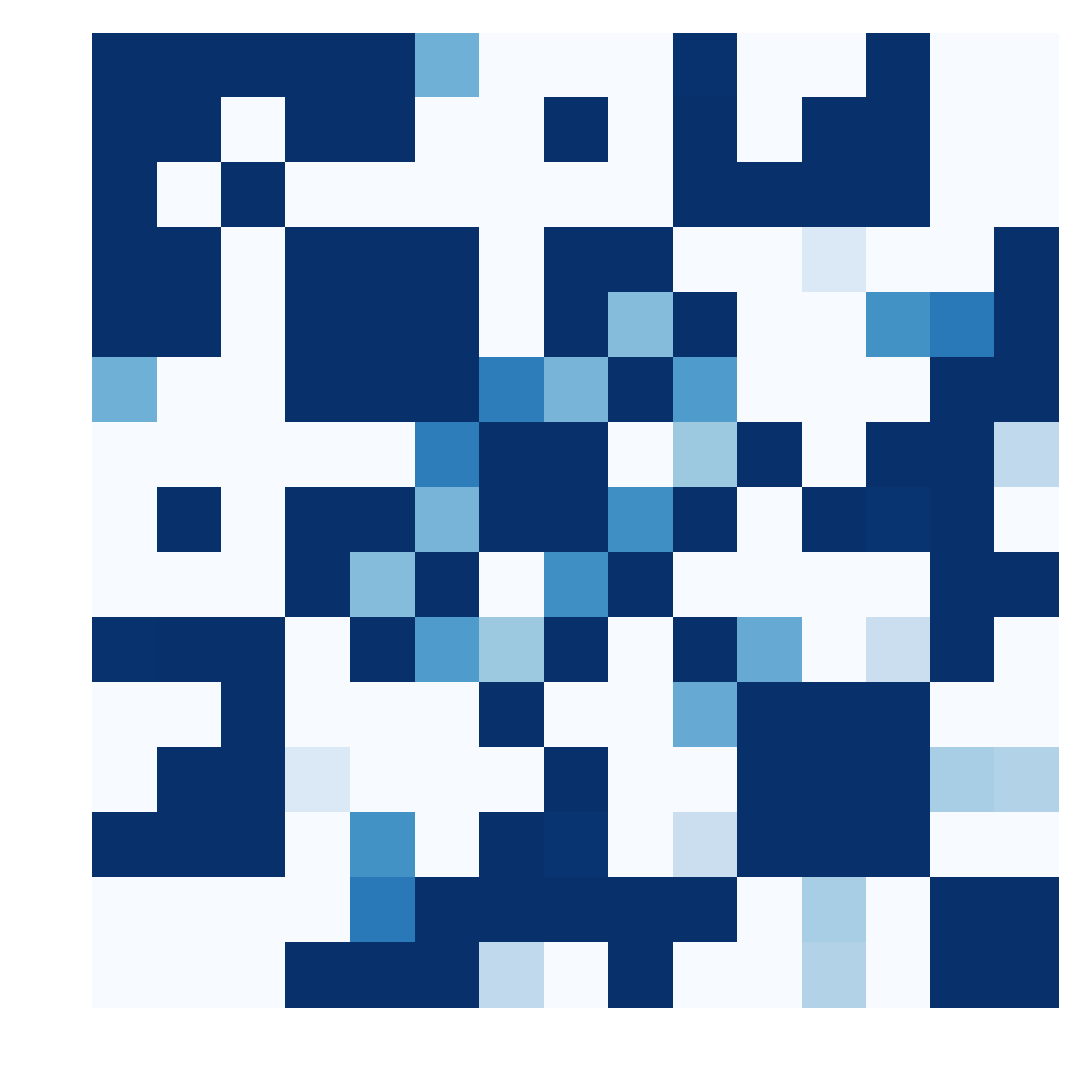}&
        \hspace{-0.4cm}\includegraphics[scale=\Scale2]{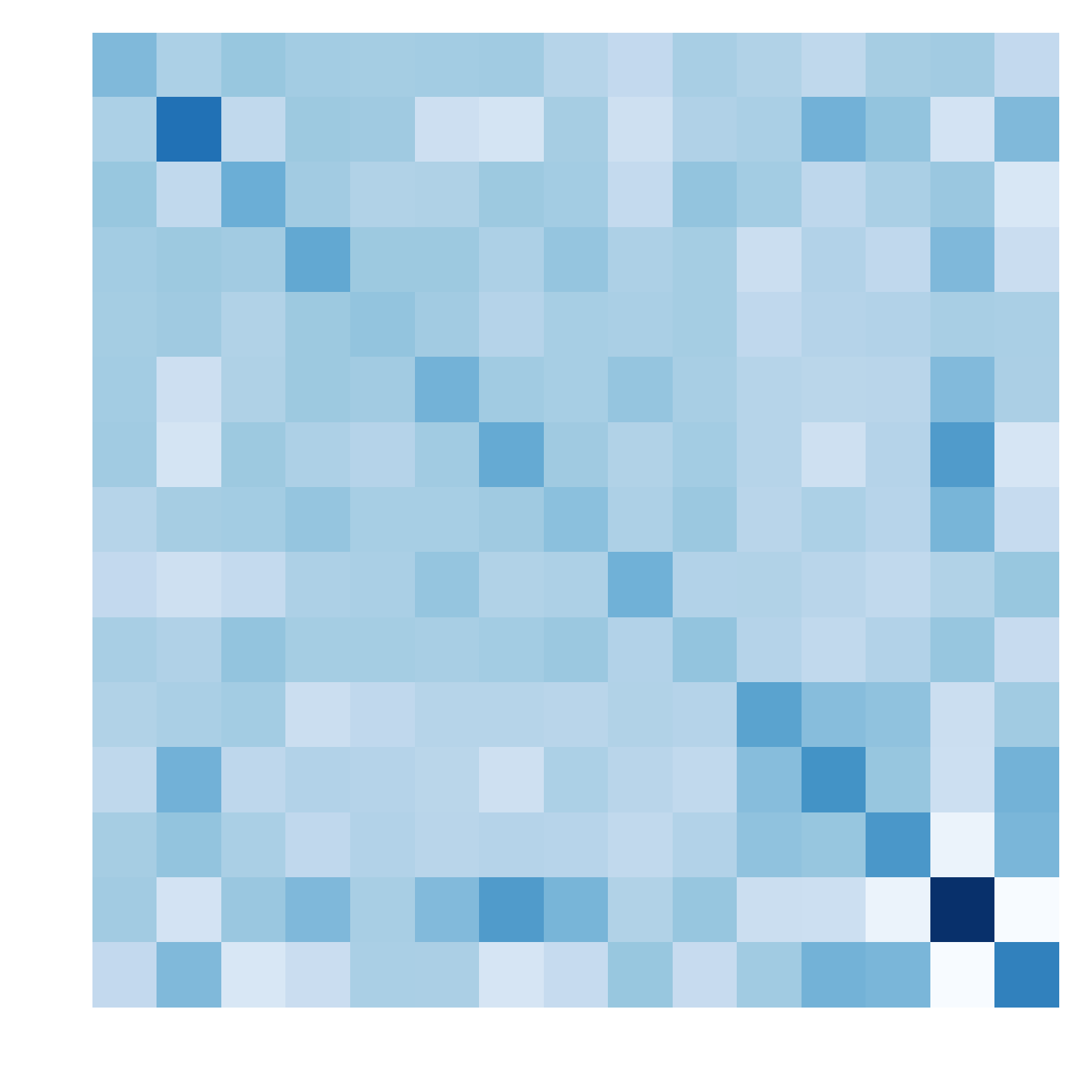}&
        \hspace{-0.4cm}\includegraphics[scale=\Scale2]{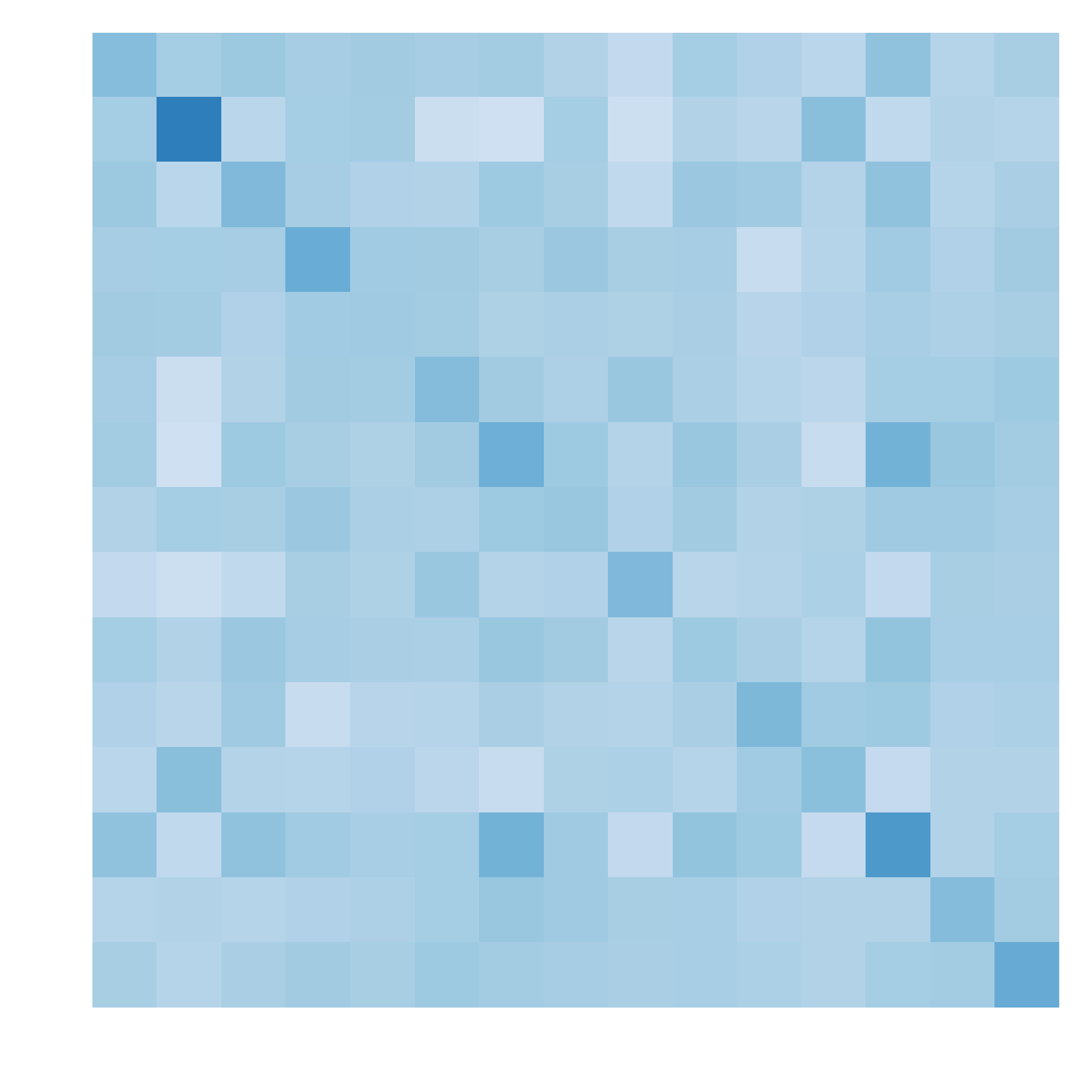}&
        \hspace{-0.35cm}\includegraphics[scale=\Scale2]{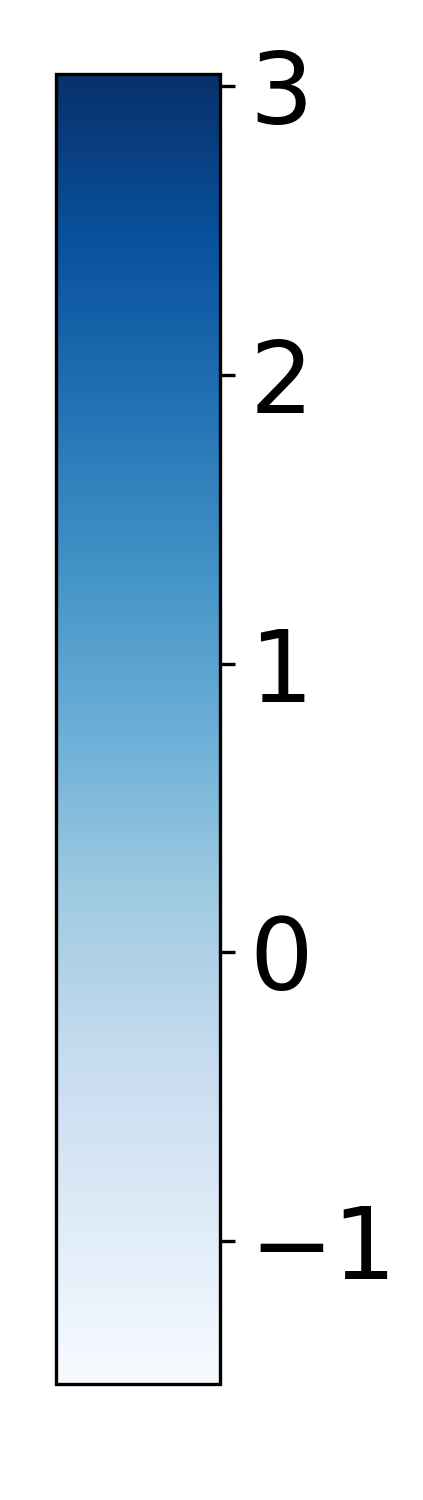} \\
        &\hspace{-0.5cm}(a) Ground-Truth & \hspace{-0.3cm}(b) Std-MT & \hspace{-0.3cm}(c) $\Sigma$-MT & \hspace{-0.4cm}(d) $\sim$Cool-MT & \hspace{-0.3cm}(e) HCool-MT & \\
    \end{tabular}
    \caption{Estimated intertask, $\bC_{1:T,1:T}$, and noise, $\bSigma_{1:T,1:T}$, covariance matrices vs. true ones (Ground-Truth)  when R=10. Std-MT and $\Sigma$-MT were trained for $P=R=10$.}
    \label{fig:synth-matrices}
\end{figure}

Figure \ref{fig:synth-boxplot} shows that $\Sigma$-MT has the highest sensitivity to the choice of $P$, and its performance degrades when the scenario complexity (defined by matrix rank $R$) increases. Std-MT shows more robustness with respect to both the selection of $P$ and the value of $R$. As was expected, both models show their best performances when $P=R$, and thus a cross validation of $P$ is paramount for these methods to perform optimally.
This sensitivity depends on the number of the parameters to be inferred, which is $TP+1$ in the case of Std-MT and $2TP$ for $\Sigma$-MT, while it is only $2T$ for both Cool-MTs. This allows our model to perform closer to the ground truth independently of the scenario complexity $R$.  

Experimentally, it can be seen in Figure \ref{fig:synth-matrices} that, while all models are capable of inferring the intertask covariance, the noise matrix is not properly inferred by the $\Sigma$-MT when scenario complexity ($R$ value) is high. Besides, comparing both versions of the Cool-MT model, the hierarchical approach is unsurprisingly slightly better at estimating the true parameters and reconstructing the noise matrix, leading to a higher consistency in its predictions as was already shown in Figure \ref{fig:synth-boxplot}.

\def\Scale2{0.15}
\begin{figure}[t!]
\centering
\footnotesize
\begin{tabular}{cccccccc}
 \rotatebox{90}{\hspace{0.5cm}$\bC_{1:T,1:T}$} & \hspace{-0.3cm}\includegraphics[scale=\Scale2]{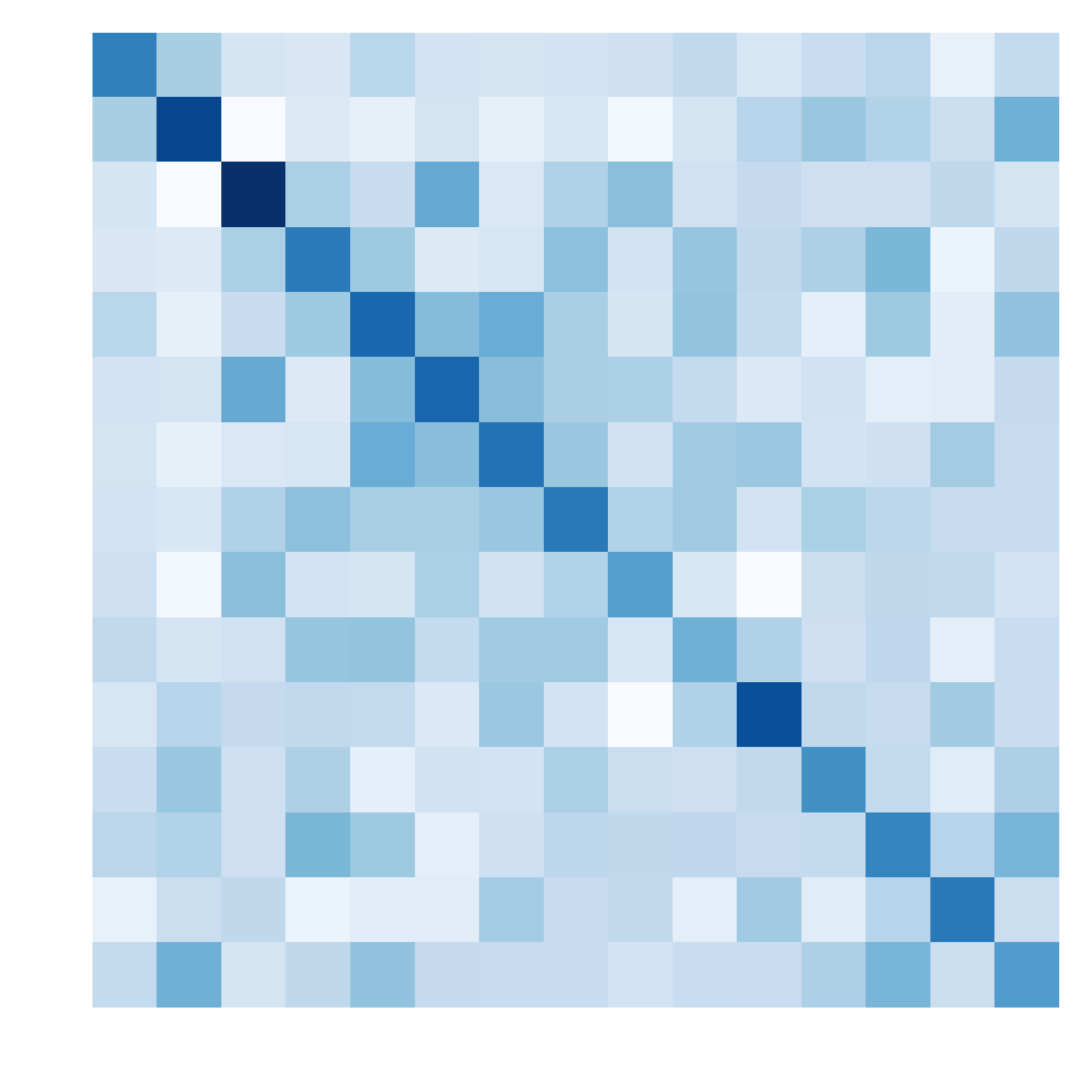}&
        \hspace{-0.3cm}\includegraphics[scale=\Scale2]{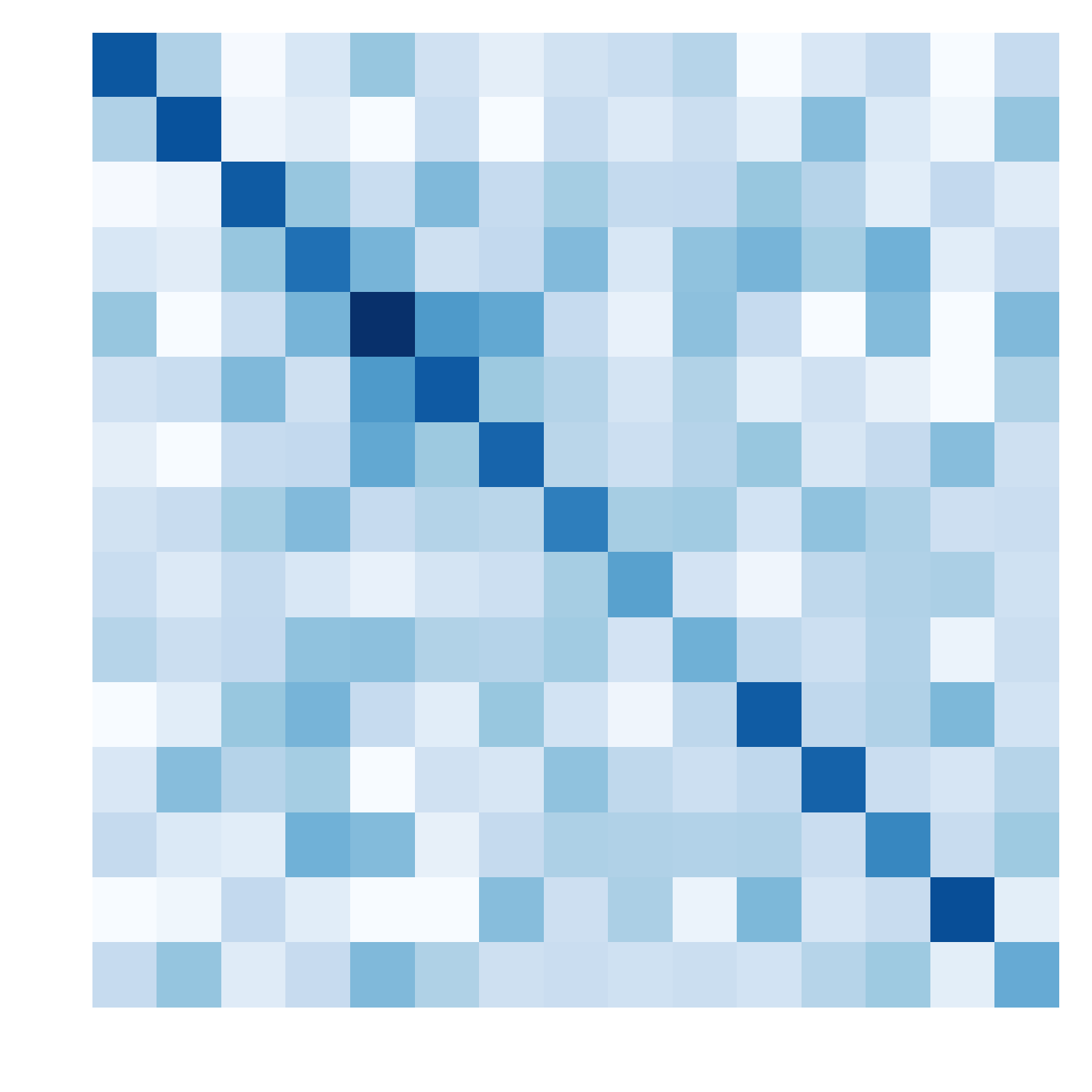}&
        \hspace{-0.3cm}\includegraphics[scale=\Scale2]{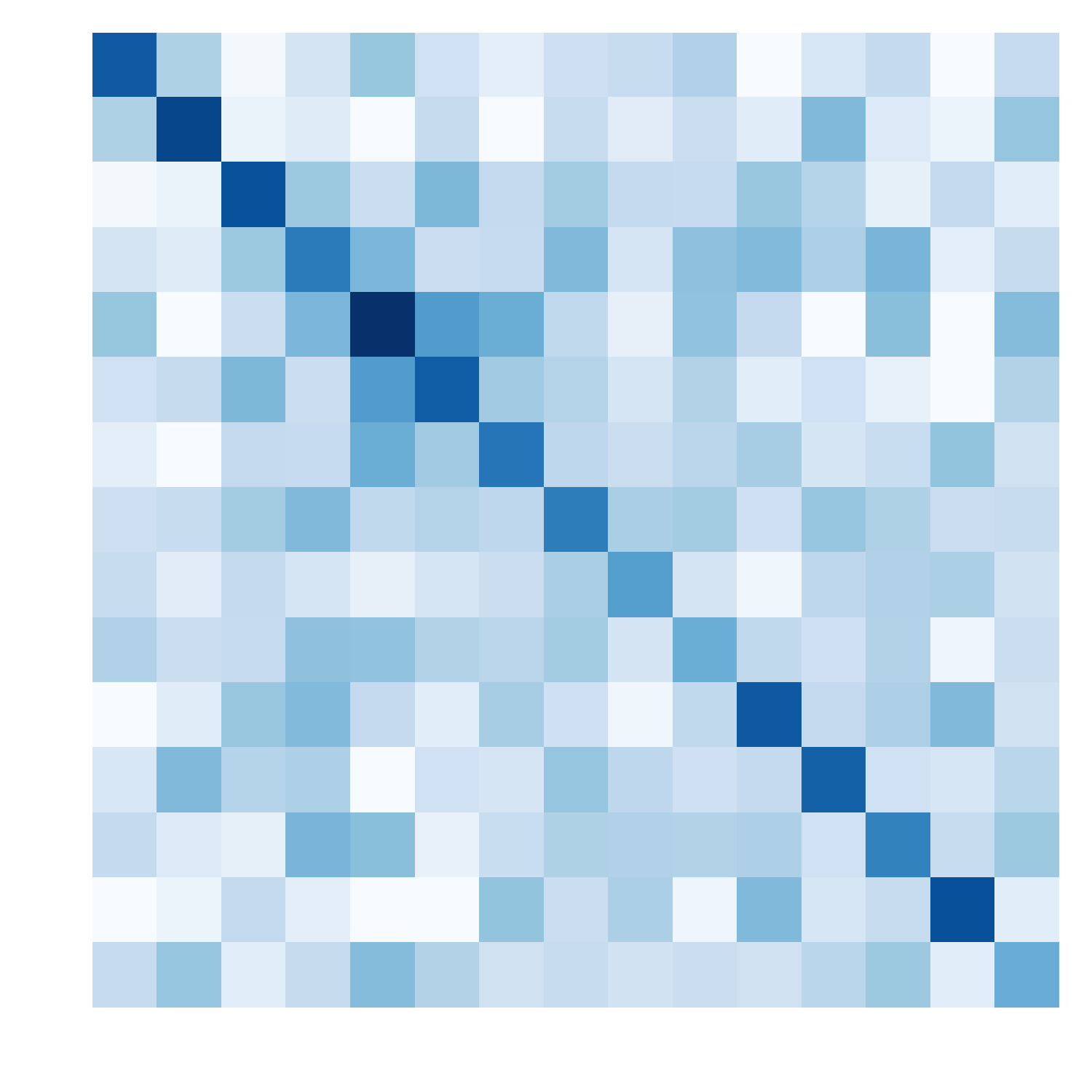}&
        \hspace{-0.3cm}\includegraphics[scale=\Scale2]{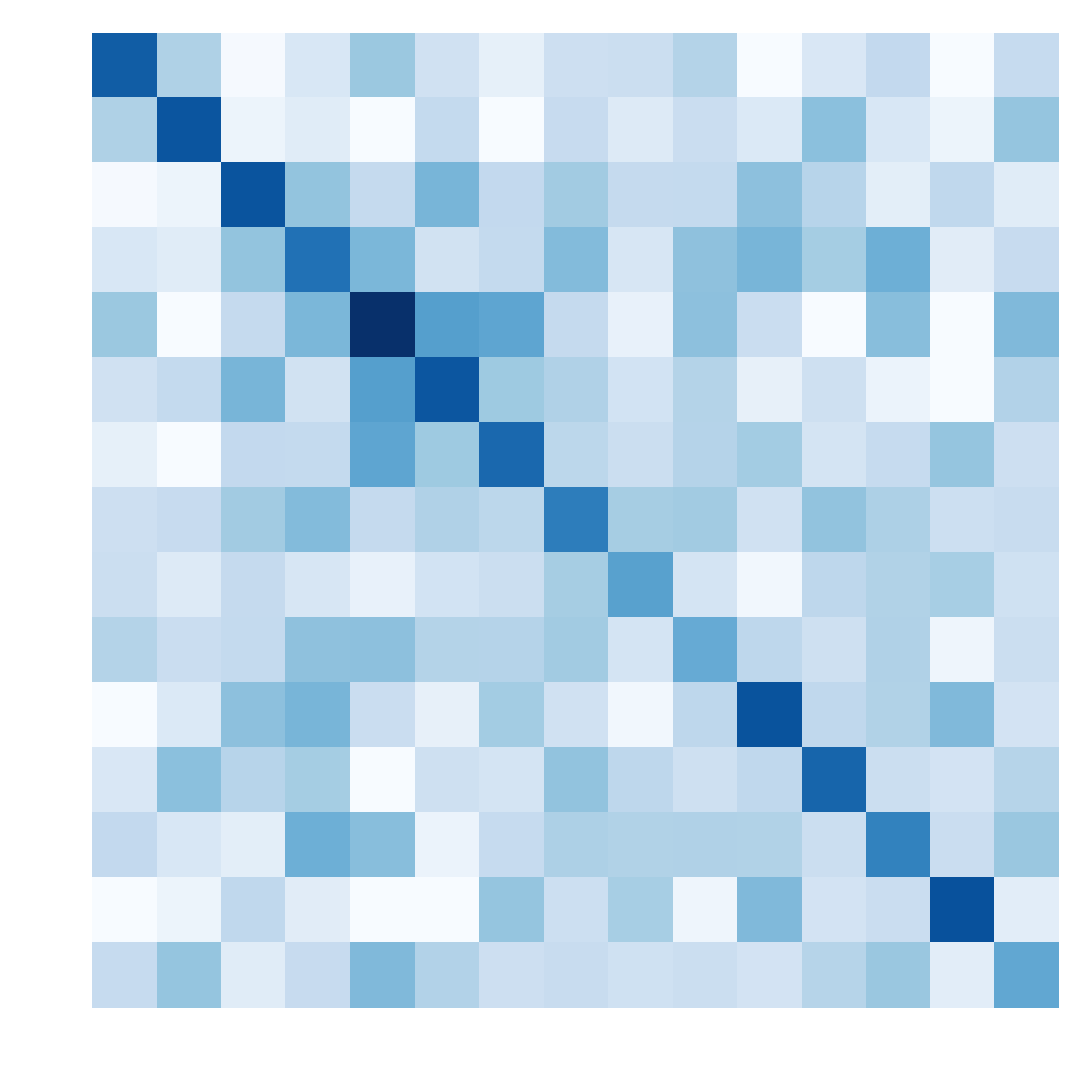}&
        \hspace{-0.3cm}\includegraphics[scale=\Scale2]{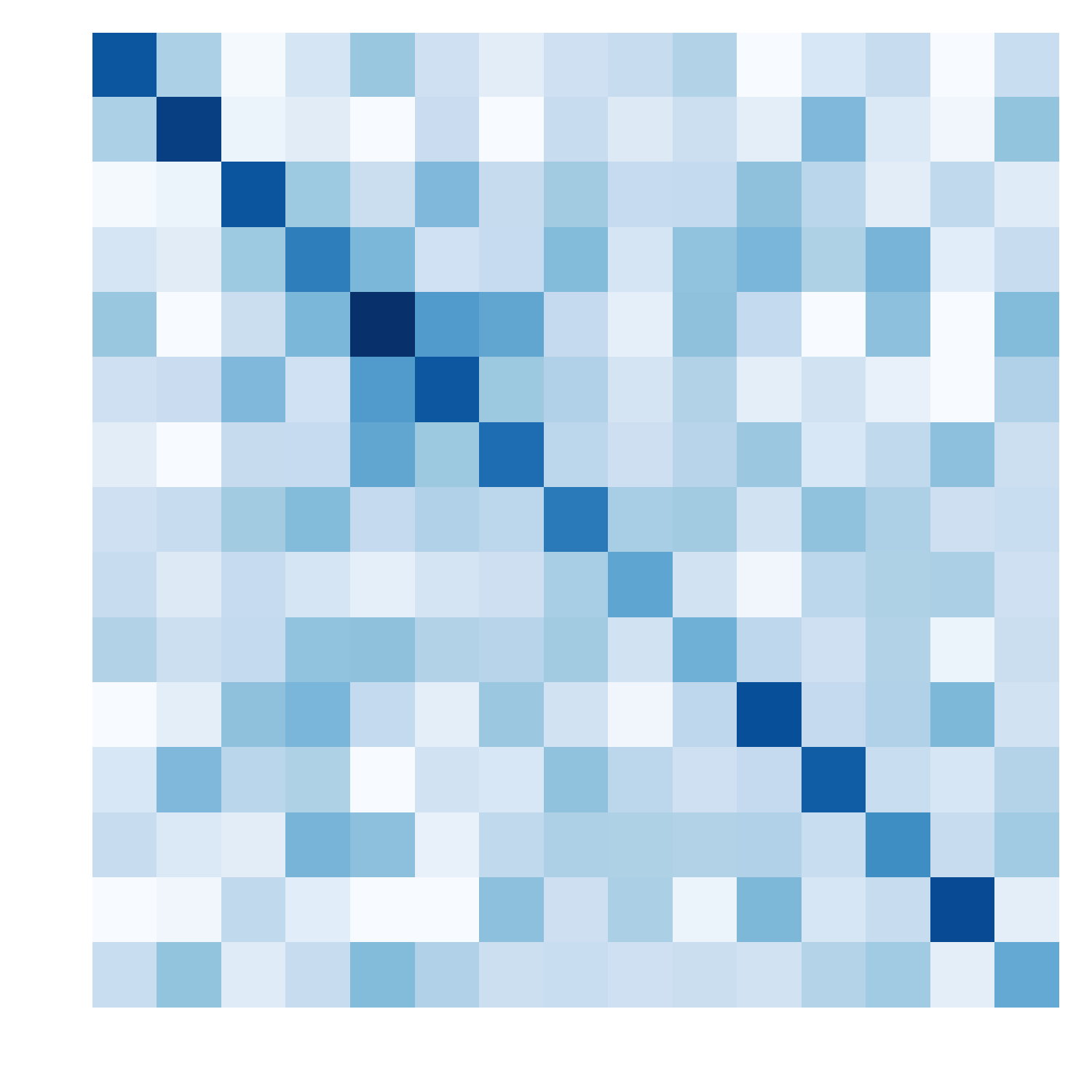}& 
        \hspace{-0.3cm}\includegraphics[scale=\Scale2]{figures/colorbar_C.png} \\
        \rotatebox{90}{\hspace{0.5cm}$\bSigma_{1:T.1:T}$}& \hspace{-0.3cm}\includegraphics[scale=\Scale2]{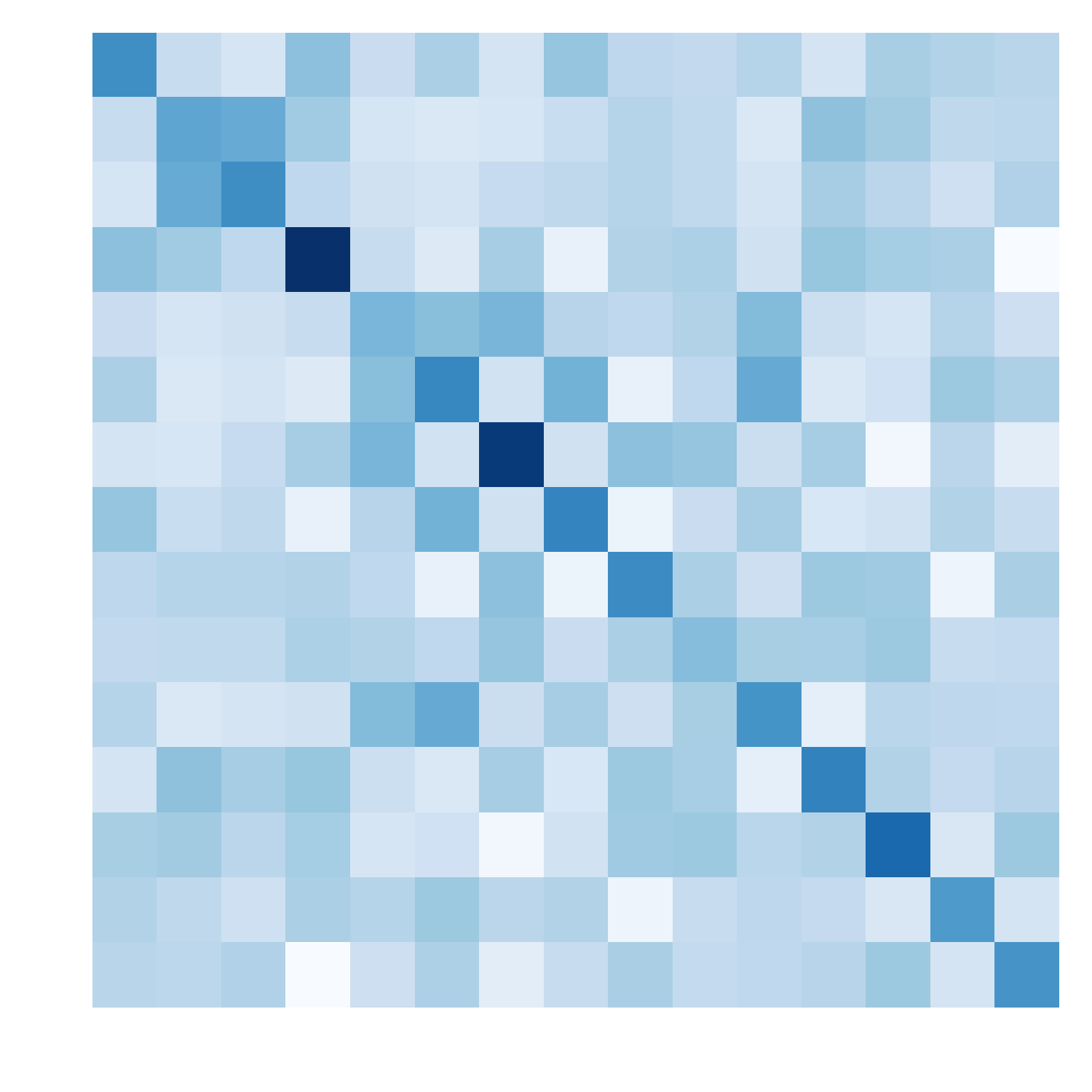}& \hspace{-0.3cm}\includegraphics[scale=\Scale2]{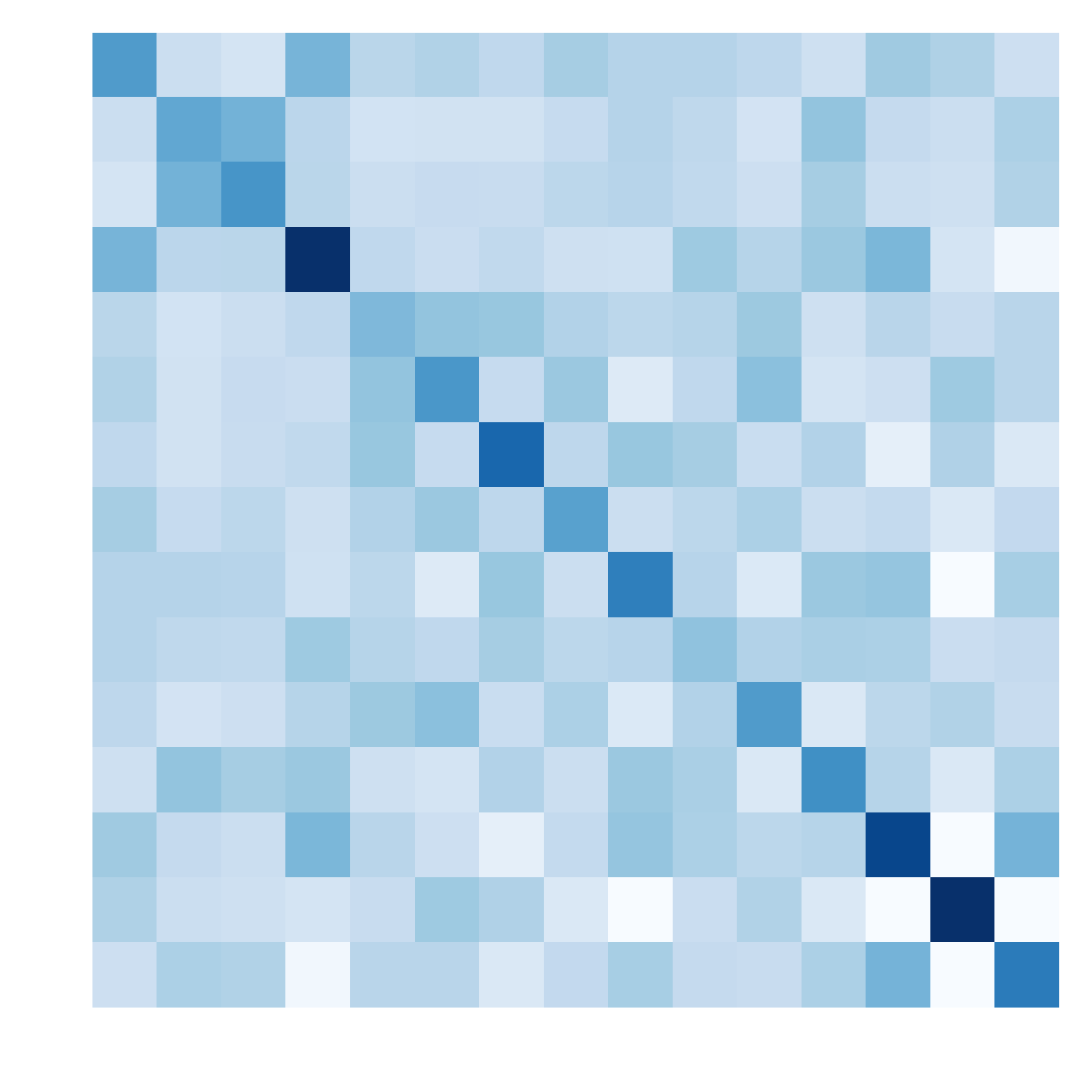}&
        \hspace{-0.3cm}\includegraphics[scale=\Scale2]{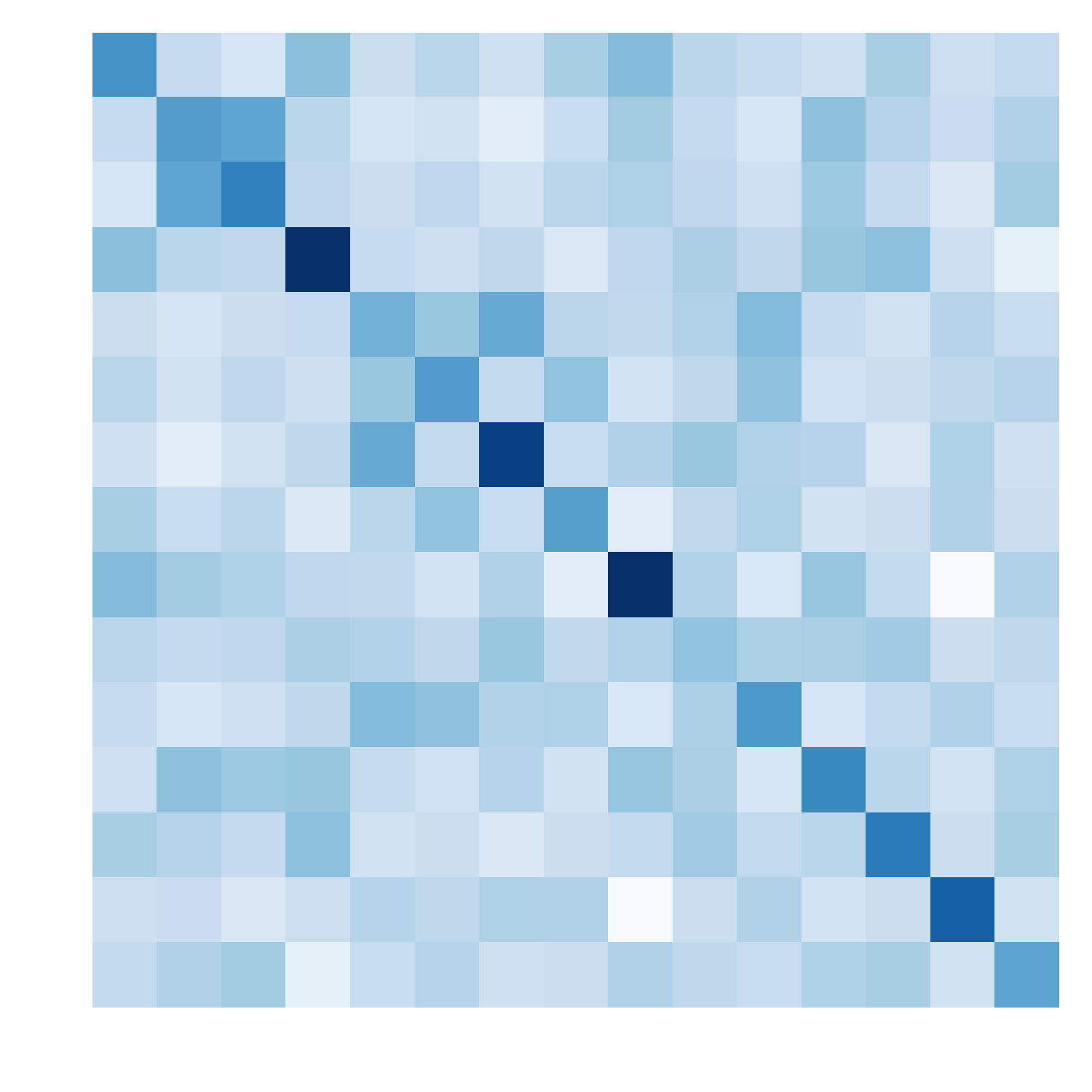}&
        \hspace{-0.3cm}\includegraphics[scale=\Scale2]{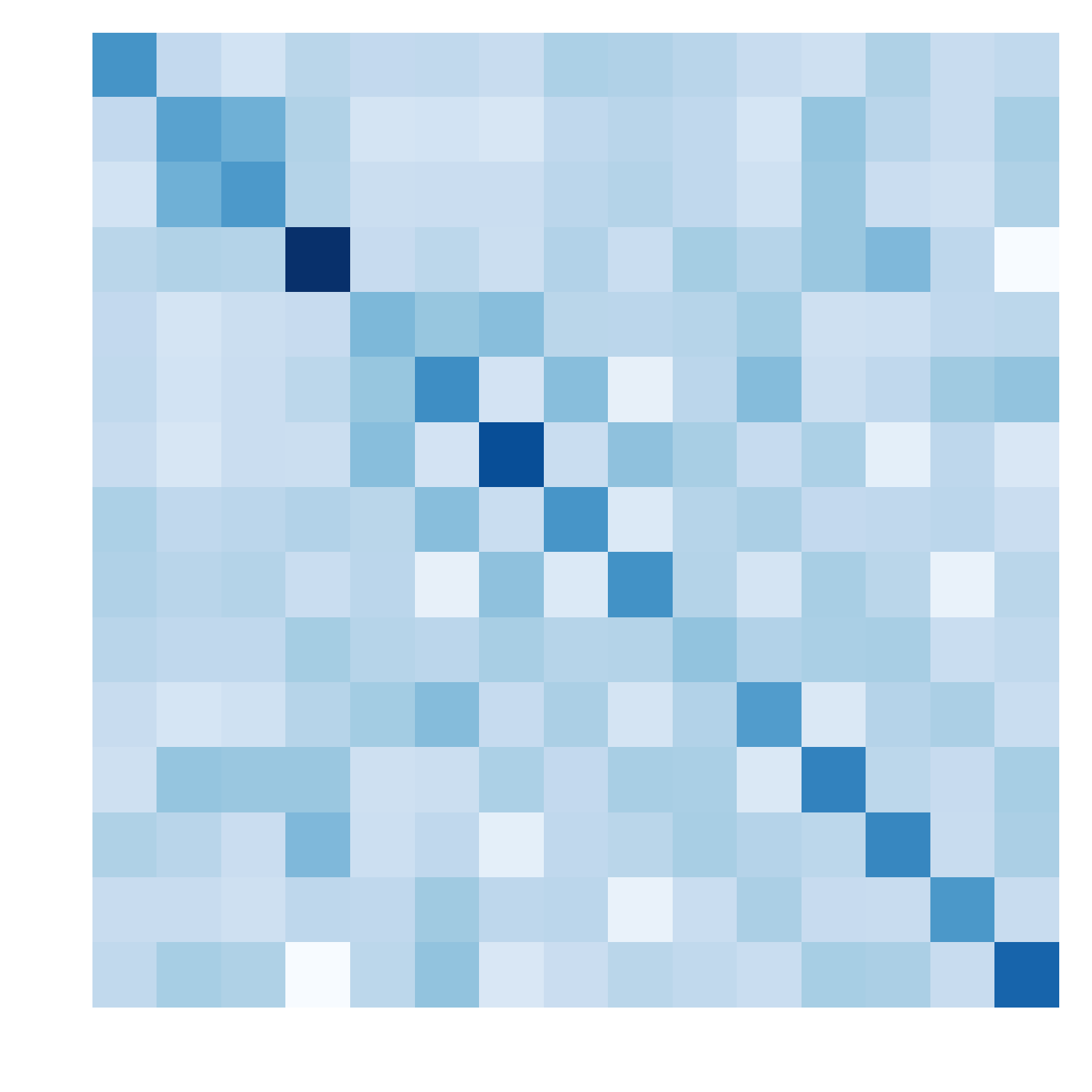}&
        \hspace{-0.3cm}\includegraphics[scale=\Scale2]{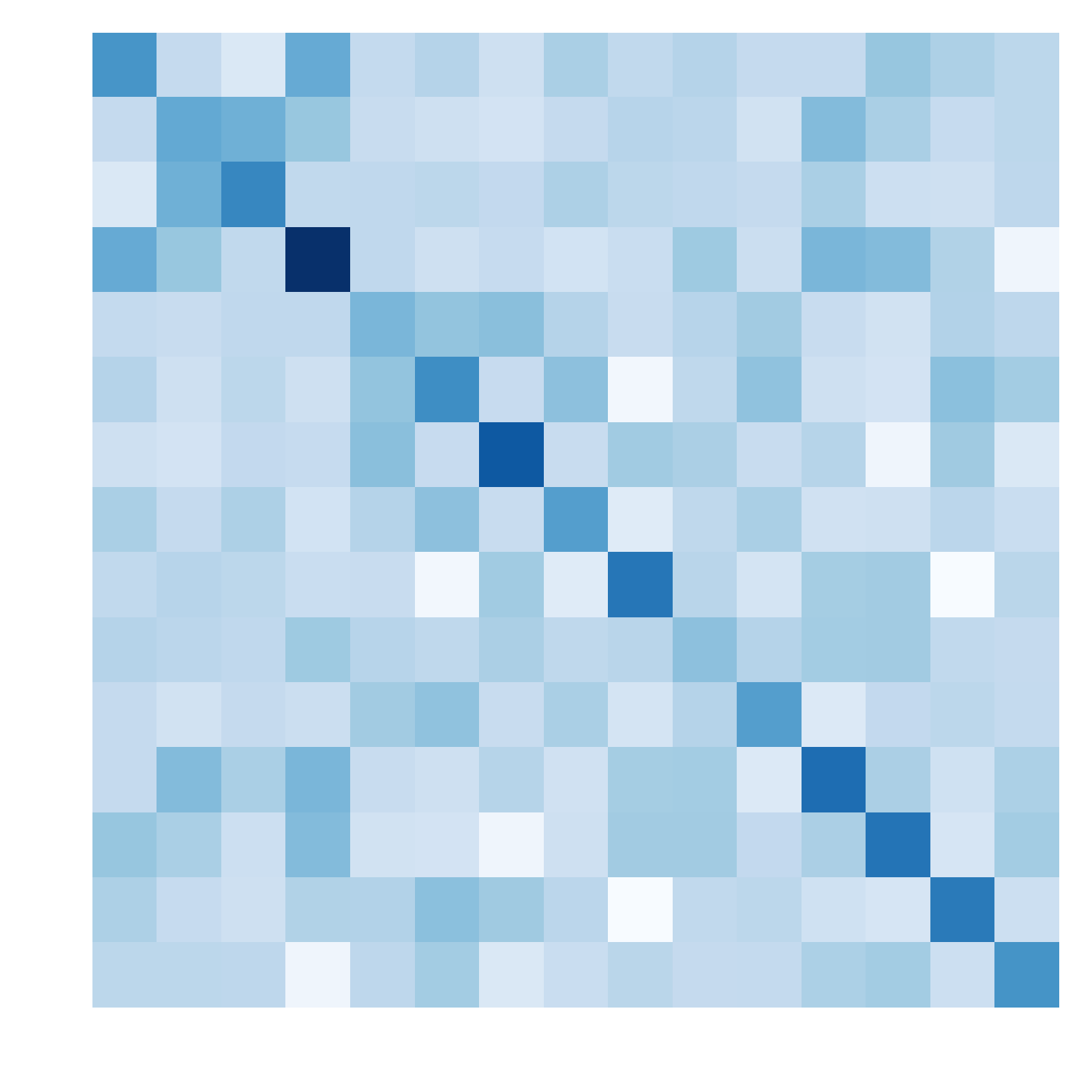} & 
        \hspace{-0.35cm}\includegraphics[scale=\Scale2]{figures/colorbar_Sigma.png} \\
        & (a) Real &(b) Random 1 & (c) Random 2 & (d) Random 3 & (e) Random 4 & \\
    \end{tabular}
    \caption{Estimated $\bC_{1:T,1:T}$ and $\bSigma_{1:T,1:T}$ covariances for four random permutations of the task order for 15 tasks with R=15.}
    \label{fig:synth-matrices_permutation}
\end{figure}

Finally we address the possibility that the model is sensitive to task ordering during the training of the Cool-MTs. As can be seen in Figure \ref{fig:synth-matrices_permutation}, the model is also consistent with regard to the order of the tasks.

\subsection{Predictive variance assessment}

One of the most important benefits of using a GP model is the ability to obtain the confidence for each prediction from the predictive distribution as well as the predicted value. In this section we study the integrity of these predictive distributions for the different MTGP models under study.

For this purpose, we have generated a simple benchmark in which a synthetic dataset is generated with the following likelihood
\begin{equation}
    p(\by) = \mathcal{N}\left(\by\left| \begin{pmatrix} \cos(2\pi x ) \\ \sin(2\pi x) \end{pmatrix} \right., \begin{pmatrix}0.1 & 0.05 \\
    0.05 & 0.1\end{pmatrix}\right)
\end{equation}
where $x\in (0, 1)$. Specifically, we have generated a training set in which the input data is clustered into two groups to force the presence of areas of lower training sample density.

For this study we have considered the Std-MTGP, the convolutional model (Conv-MT) proposed in \cite{GPflow2017} and the two proposed versions of the Cool-MT. The $\Sigma$-MT model was left out due to the lack of a non-linear kernel implementation. For the reference models (Std-MT and Conv-MT), we have found that their libraries only provide an estimation of the confidence interval task-by-task. Therefore, we start by analyzing these intervals. As can be seen in Figure \ref{fig:confidence_interval}, we observe that all models arrive at similar task-wise confidence intervals. 

\begin{figure}[!ht]
    \centering
    \begin{tabular}{c}
         \includegraphics[width=0.8\textwidth]{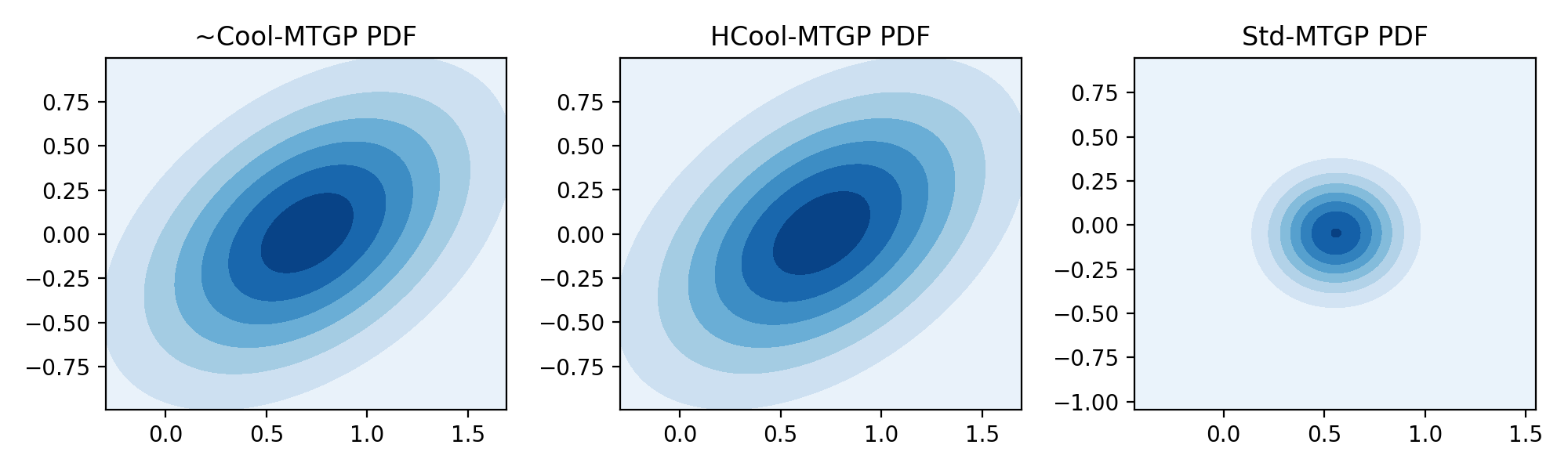} \\
         \includegraphics[width=0.8\textwidth]{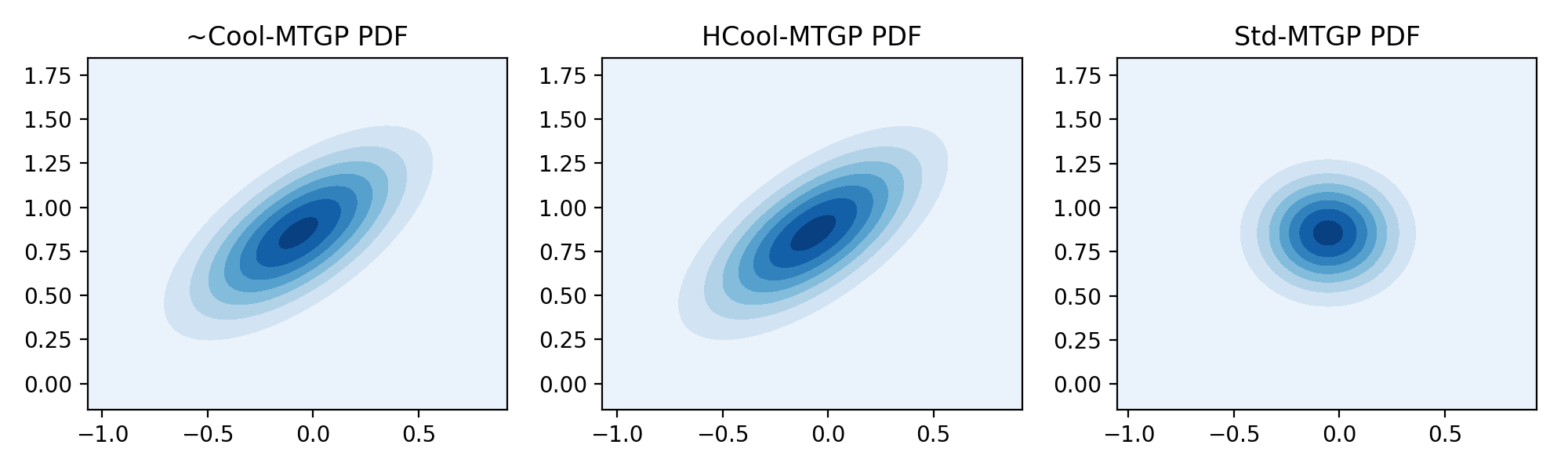} \\
         \includegraphics[width=0.8\textwidth]{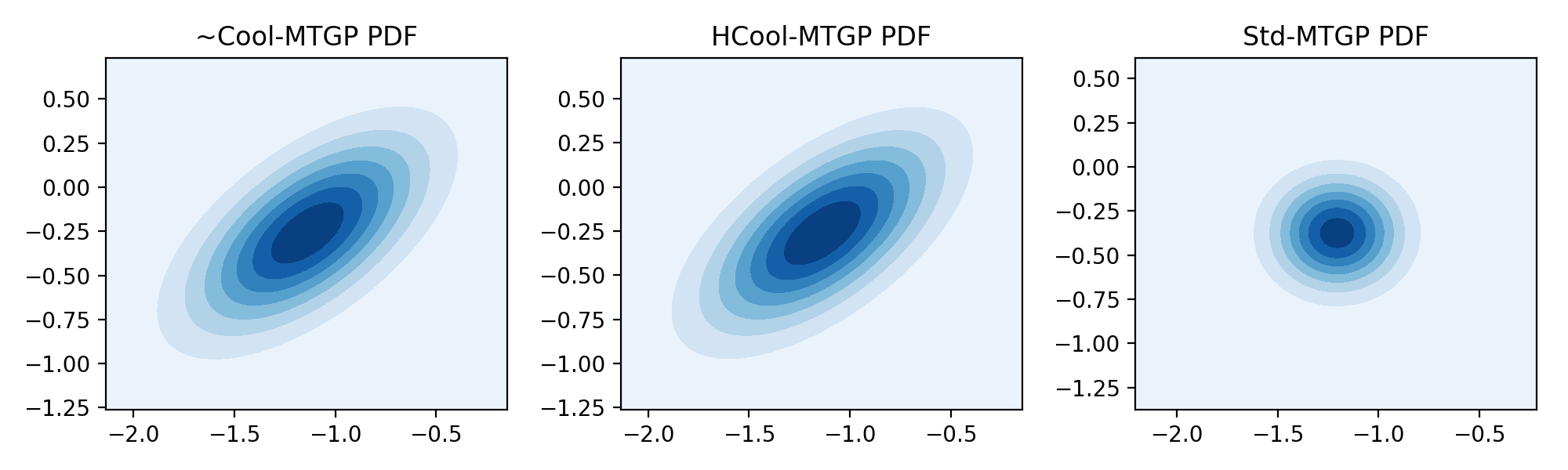} \\
    \end{tabular}
    \caption{\review{Level curves of the joint predictive distribution in the output space, represented by  $f_1(x^*)$ in the horizontal axis and   $f_2(x^*)$ in the vertical axis for different values of $x^*$ ($x^* = 0$ (top), $x^* = 0.25$ (center) and $x^* = 0.5$ (bottom))} \label{fig:predictive_dist}}
\end{figure}

Going deeper into this analysis, we have recovered the complete predictive distribution for the Cool-MT and for the Std-MT (this can be easily done using their learnt $C_{1:T,1:T}$ and $\Sigma_{1:T,1:T}$ matrices); however, we have not been able to obtain this distribution for the Conv-MT due to the complexity of this model and the black-box nature of its implementation, which has made it difficult to recover the hyperparameters. To analyse the complete predictive posterior for these models we have selected three test samples: $x^* = 0$, $x^* = 0.25$ and $x^* = 0.5$. This way we cover regions with high and low predictive confidence. The results, depicted in Figure \ref{fig:predictive_dist} allow us to appreciate clear differences:

\begin{itemize}
    \item While Std-MT does offer a full predictive distribution, it considers the noise to be independent among tasks. This prevents the method from modelling the relationships among the tasks in the predictive distribution. Therefore its predictive distribution covariance matrix tends to be diagonal. 
    
    \item Both versions of Cool-MT produce predictive posteriors that adequately model the correlation among tasks with a full covariance matrix. 
    
    \item In regions where the confidence is lower, the distribution for the Cool-MT model widens yet it adequately retains its shape. This isn't the case for Std-MT, which seems to be insensitive to the level of confidence for a particular region.
    
    \item There are no appreciable differences between the predictive distributions obtained by HCool-MT and $\sim$Cool-MT.
\end{itemize}

\def\Scale2{0.45}

\begin{figure}[!ht]
    \centering
    \begin{tabular}{c}
    \hspace{-1cm}
         \includegraphics[width=\Scale2\textwidth]{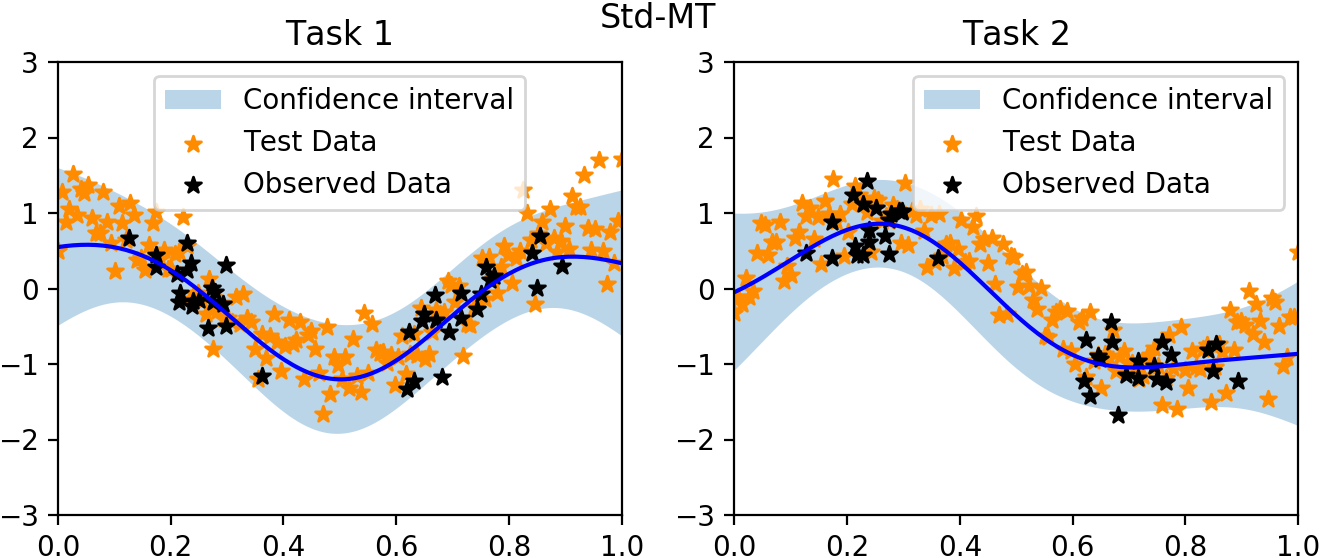}
         \includegraphics[width=\Scale2\textwidth]{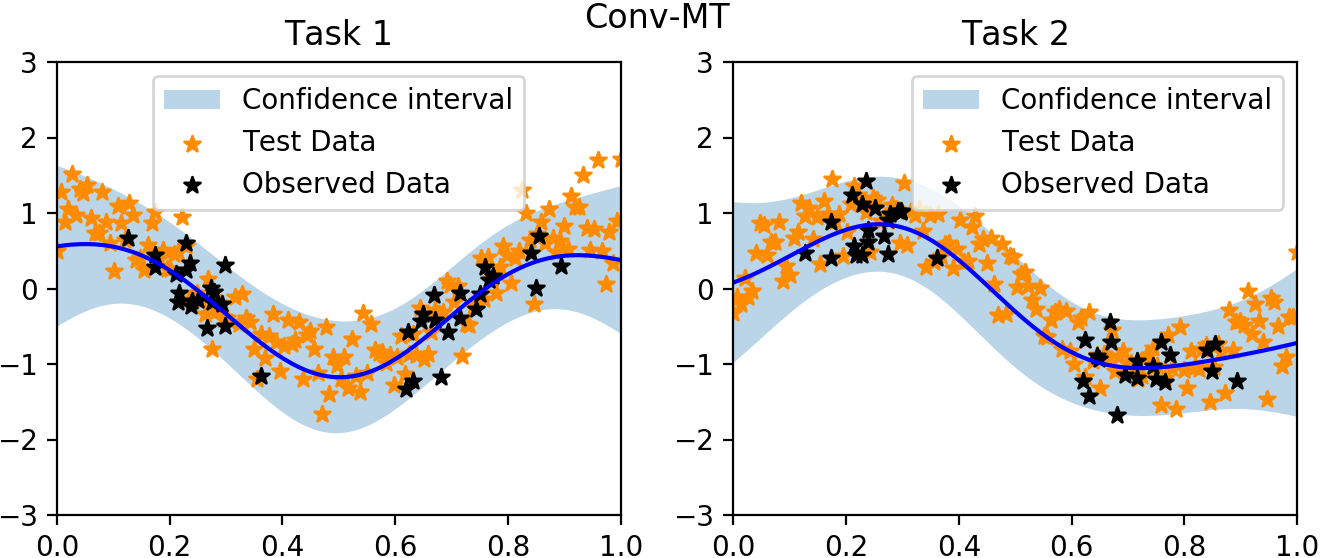} \\
          \hspace{-1cm}
         \includegraphics[width=\Scale2\textwidth]{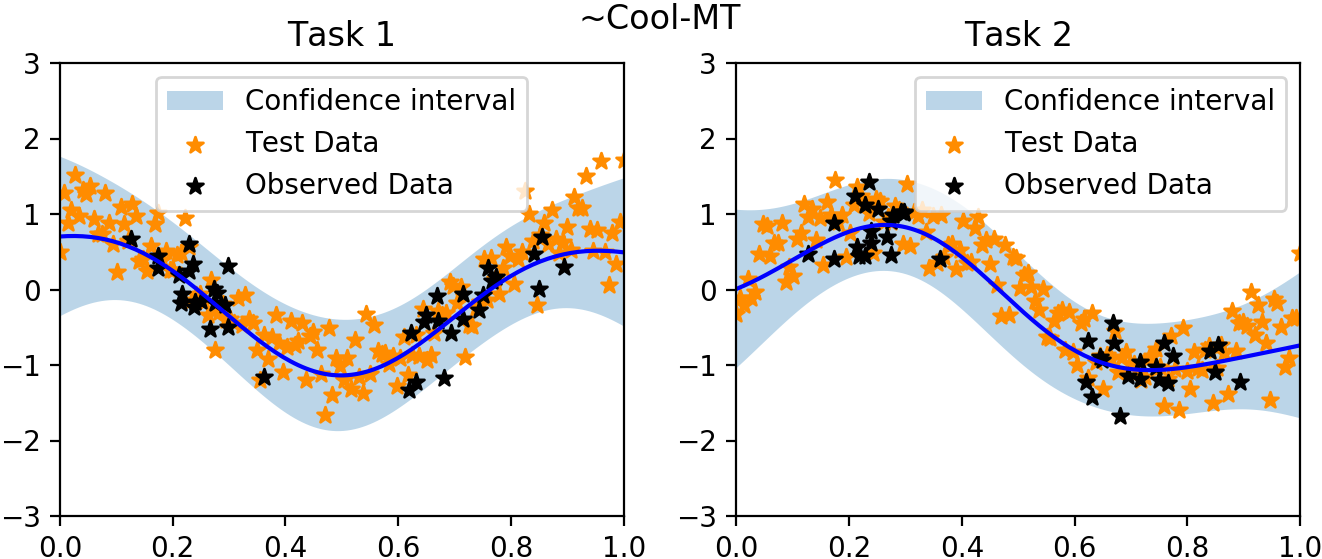} 
         \includegraphics[width=\Scale2\textwidth]{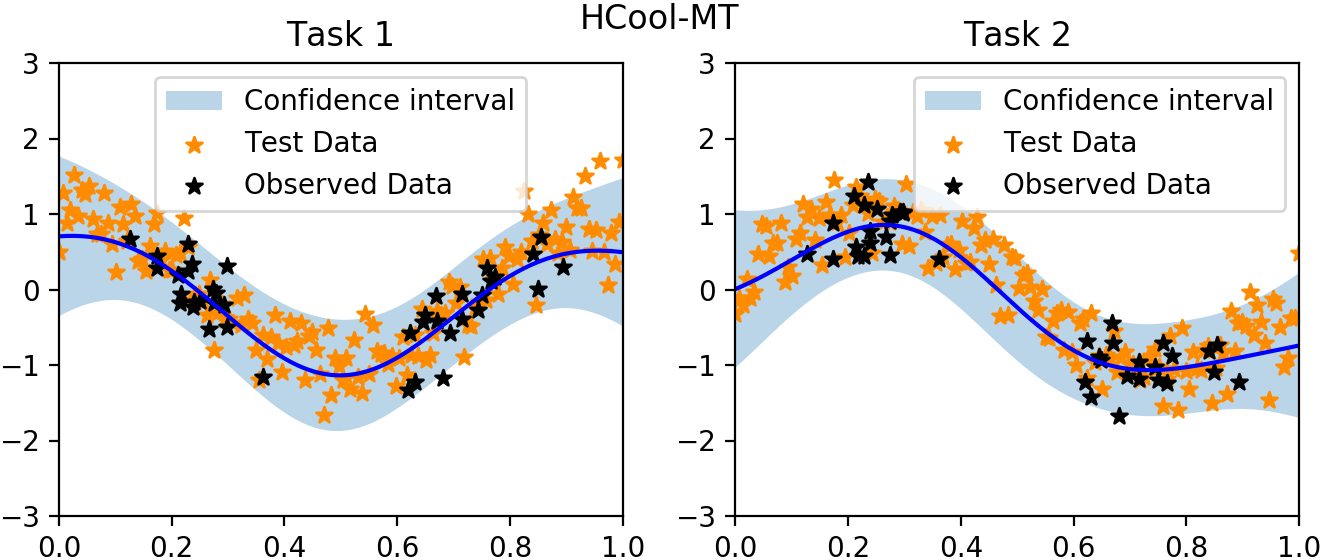} \\
    \end{tabular}
    \caption{\review{Predictive mean and confidence interval (vertical axis) with respect to the different input values (horizontal axis) for both tasks and over the different models under study.} \label{fig:confidence_interval}}
\end{figure}

\subsection{Real data benchmarks}

In this experiment we use ten real world scenarios to compare the Cool-MT model's capabilities to those of the Std-MT and Conv-MT models\footnote{We have been unable to include $\Sigma$-MT in this part of our study due to convergence issues with the available implementation, as well as the lack of a non-linear kernel version.}.

To accomplish this, we have made use of the collection of datasets featured in \cite{spyromitros2016multi}, which can be found online at \url{http://mulan.sourceforge.net/}. These datasets offer a good variety in sample size, input dimensionality and number of tasks, as can be seen in Table \ref{tab: datasets}. Due to the small performance differences seen in the previous section between the hierarchical and approximate versions of the Cool-MT, for the real world benchmarks we have only used the approximate model.

\begin{table}[h!]
\centering
\caption{\review{Real world datasets used in this work.}\label{tab: datasets}}
\footnotesize
\begin{tabular}{lrrr}
\toprule
\multicolumn{1}{c}{\textbf{Dataset}} & \multicolumn{1}{c}{\textbf{Samples}} & \multicolumn{1}{c}{\textbf{Features}} & \multicolumn{1}{c}{\textbf{Tasks}} \\
\midrule
andro & 49 & 30 & 6 \\
enb & 768 & 8 & 2 \\
edm & 154 & 16 & 2 \\
slump & 103 & 7 & 3 \\
oes10 & 403 & 298 & 16 \\
oes97 & 334 & 263 & 16 \\
atp1d & 337 & 411 & 6 \\
atp7d & 296 & 411 & 6 \\
scm1d & 9803 & 280 & 16 \\
scm20d & 8966 & 61 & 16 \\
\bottomrule
\addlinespace[10pt]                             
\end{tabular}

\end{table}

In all cases a standard normalization of the data was applied to both the input variables and output targets. The normalization parameters were obtained using the training partition only.  Ten iterations were run with a random 80\%/20\% training/test partitioning of the data. In larger datasets (\# samples $>$ 1000) we dedicated 800 samples to the training set and 200 to the test set, limiting the number of total samples. All models were trained with both a linear kernel and a squared exponential (SE) kernel.

Regarding to hyperparameter setting, Cool-MT doesn't need validation of any hyperparameters since they are all inferred by the algorithm. The kernel objects require the definition of an exploration region for their relevant parameters, so a sufficiently wide value range of $(10^{-10}, 10^{3})$ was used to ensure good convergence of the optimizers. The Std-MT model requires the selection of hyperparameter $P$; for this setting, after trying values $1$, $T/2$ and $T$ on a few datasets (where $T$ is the number of tasks) using a small validation partition, it became apparent that the best value was consistently $P=1$; and the kernel parameters were initialised using their default settings. The Conv-MT model requires the user to set the number of inducing points. After a brief exploration we settled on 50\% of the size of the training partition, achieving a good performance while avoiding convergence issues.

\begin{table}[h!]
\centering
\caption{\review{Real dataset benchmark results using a linear kernel: RMSE averaged over tasks. Best results in bold.}\label{tab:real-results-linear}}
\footnotesize
\begin{tabular}{lrrrr}
\toprule
\multicolumn{1}{c}{\textbf{Dataset}} & \multicolumn{1}{c}{\textbf{Indep. GPs}} & \multicolumn{1}{c}{\textbf{Std-MT}} & \multicolumn{1}{c}{\textbf{Conv-MT}} & \multicolumn{1}{c}{\textbf{Cool-MT}} \\
\midrule
\textbf{andro} & $\mathbf{0.74 \pm 0.14}$ & $0.81 \pm 0.15$ & $0.77 \pm 0.10$ & $0.75 \pm 0.10$ \\
\textbf{enb} & $\mathbf{0.31 \pm 0.02}$ & $\mathbf{0.31 \pm 0.03}$ & $\mathbf{0.31 \pm 0.02}$ & $\mathbf{0.31 \pm 0.02}$ \\
\textbf{edm} & $0.79 \pm 0.05$ & $\mathbf{0.78 \pm 0.05}$ & $\mathbf{0.78 \pm 0.05}$ & $\mathbf{0.78 \pm 0.05}$ \\
\textbf{slump} & $0.68 \pm 0.07$ & $0.68 \pm 0.07$ & $0.68 \pm 0.07$ & $\mathbf{0.67 \pm 0.07}$ \\
\textbf{oes10} & $0.47 \pm 0.17$ & $0.38 \pm 0.14$ & $\mathbf{0.35 \pm 0.13}$ & $\mathbf{0.35 \pm 0.13}$ \\
\textbf{oes97} & $0.56 \pm 0.18$ & $0.41 \pm 0.12$ & $\mathbf{0.39 \pm 0.11}$ & $\mathbf{0.39 \pm 0.11}$ \\
\textbf{atp1d} & $0.50 \pm 0.05$ & $0.49 \pm 0.05$ & $\mathbf{0.42 \pm 0.07}$ & $\mathbf{0.42 \pm 0.07}$ \\
\textbf{atp7d} & $0.70 \pm 0.12$ & $0.64 \pm 0.11$ & $\mathbf{0.56 \pm 0.07}$ & $\mathbf{0.56 \pm 0.07}$ \\
\textbf{scm1d} & $0.29 \pm 0.02$ & $0.27 \pm 0.02$ & $\mathbf{0.24 \pm 0.02}$ & $\mathbf{0.24 \pm 0.02}$ \\
\textbf{scm20d} & $0.36 \pm 0.03$ & $0.36 \pm 0.02$ & $\mathbf{0.36 \pm 0.02}$ & $\mathbf{0.36 \pm 0.02}$ \\
\bottomrule
\addlinespace[10pt]
\end{tabular}

\end{table}

\begin{table}[h!]
\centering
\caption{\review{Real dataset benchmark results using a squared exponential kernel. RMSE averaged over tasks. Best results in bold.}\label{tab:real-results-rbf}}
\footnotesize
\begin{tabular}{lrrrr}
\toprule
\multicolumn{1}{c}{\textbf{Dataset}} & \multicolumn{1}{c}{\textbf{Indep. GPs}} & \multicolumn{1}{c}{\textbf{Std-MT}} & \multicolumn{1}{c}{\textbf{Conv-MT}} & \multicolumn{1}{c}{\textbf{Cool-MT}} \\
\midrule
\textbf{andro} & $0.62 \pm 0.10$ & $\mathbf{0.42 \pm 0.07}$ & $0.46 \pm 0.08$ & $\mathbf{0.42 \pm 0.07}$ \\
\textbf{enb} & $0.30 \pm 0.02$ & $0.15 \pm 0.02$ & $0.16 \pm 0.02$ & $\mathbf{0.13 \pm 0.02}$ \\
\textbf{edm} & $0.70 \pm 0.06$ & $0.73 \pm 0.06$ & $\mathbf{0.71 \pm 0.06}$ & $0.72 \pm 0.05$ \\
\textbf{slump} & $0.68 \pm 0.11$ & $0.67 \pm 0.07$ & $\mathbf{0.61 \pm 0.08}$ & $0.63 \pm 0.07$ \\
\textbf{oes10} & $0.76 \pm 0.51$ & $0.85 \pm 0.45$ & $1.03 \pm 0.43$ & $\mathbf{0.57 \pm 0.41}$ \\
\textbf{oes97} & $0.80 \pm 0.56$ & $0.81 \pm 0.51$ & $0.99 \pm 0.49$ & $\mathbf{0.63 \pm 0.46}$ \\
\textbf{atp1d} & $0.49 \pm 0.10$ & $0.81 \pm 0.12$ & $0.90 \pm 0.12$ & $\mathbf{0.41 \pm 0.07}$ \\
\textbf{atp7d} & $0.94 \pm 0.15$ & $0.88 \pm 0.14$ & $0.94 \pm 0.15$ & $\mathbf{0.56 \pm 0.10}$ \\
\textbf{scm1d} & $0.26 \pm 0.03$ & $0.23 \pm 0.02$ & $0.99 \pm 0.06$ & $\mathbf{0.22 \pm 0.02}$ \\
\textbf{scm20d} & $0.33 \pm 0.03$ & $\mathbf{0.27 \pm 0.03}$ & $0.52 \pm 0.28$ & $0.28 \pm 0.03$ \\
\bottomrule
\addlinespace[10pt]
\end{tabular}

\end{table}

Tables \ref{tab:real-results-linear} and \ref{tab:real-results-rbf} show the benchmark results in terms of the root mean squared error (RMSE) for the linear and SE kernels respectively. In the linear case all models perform similarly, with a slight advantage in favour of both the Conv-MT and Cool-MT. A strong improvement in performance is obtained in all cases using  a non-linear kernel, where the Cool-MT comes clearly on top in most datasets. After analysing the values for the SE kernel length-scale parameter learnt by all the models, it is clear that Conv-MT and, in some cases, Std-MT are unable to achieve a correct estimation. We believe that this is due to our model's reduced number of parameters to be learnt, making its adjustment easier.

\review{The linear algorithms show a low performance in the dataset \textbf{andro} (Table \ref{tab:real-results-linear}), but slightly better for the independent GP and the Cool-MTGP in spite of having to fit more parameters in the case of the MTGP model. However, Table \ref{tab:real-results-rbf} shows that the performance for this dataset is significantly increased if a nonlinear model is used, and we can therefore conclude that none of the linear models show a good performance. In this particular case, the Std-MTGP and the Cool MTGP show similar performances, suggesting that the problem is nonlinear and presents correlation between tasks, but the noise model of the standard multitask GP seems to be adequate.}

\subsection{Computational performance analysis}

In this last section we evaluate the computational performance of some of the methods under study when executed on a CPU and a GPU. For this purpose, we have selected the Std-MT and Conv-MT, since they are efficiently implemented over Pytorch and TensorFlow, and we have designed a wrapper over the Pytorch GP implementation for the proposed $\sim$Cool-MT approach in order to run it on GPUs. We have measured the runtime and MSE performances of each algorithm with a linear kernel for different sized ($N$) training partitions of the scm20d dataset considering only 4 tasks. Computational times are averaged over $50$ iterations. $50$ optimization iterations were used for the Std-MT and $\sim$Cool-MT methods, whereas Conv-MT needed $200$ to obtain accurate results. The experiment was carried out on an Intel Core i9 Processor using a single core (3.3GHz, 98GB RAM) and a GeForce RTX 2080Ti GPU (2944 Cuda Cores, 1.545GHz, 10.76GB VRAM).

Figure \ref{fig:time_4} shows the evolution of the runtime and MSE with $N$. Conv-MT and $\sim$Cool-MT show similar MSE, but the computational time of Conv-MT grows much faster with the number of data. We conclude that $\sim$Cool-MT presents the best trade-off between accuracy and computational burden.

Notably, while Std-MT's implementation is specific to GPUs using the optimizers provided by Pytorch, for now $\sim$Cool-MT only uses a wrapper. Despite this, $\sim$Cool-MT achieves comparable performance. Additional improvements can be expected with an implementation 
tailored for parallelization.

\def\Scale3{0.35}
\begin{figure}[t!]
\centering
\begin{tabular}{ccc}
        \hspace{-0.3cm}\includegraphics[scale=\Scale3]{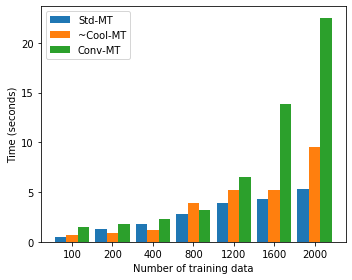}&
        \hspace{-0.3cm}\includegraphics[scale=\Scale3]{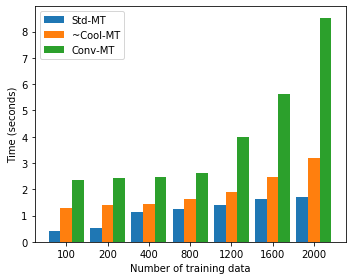}&
        \hspace{-0.3cm}\includegraphics[scale=0.63]{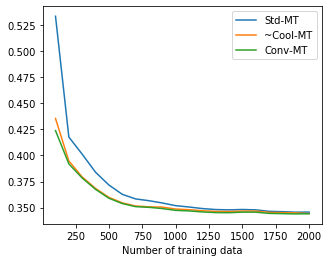}\\
        (a) CPU time & (b) GPU time & (c) MSE evolution \\
       
    \end{tabular}
    \caption{Runtime and accuracy comparison on the scm20d dataset with 4 tasks for different numbers of training samples. Conv-MT and $\sim$Cool-MT are competitive in MSE, but $\sim$Cool-MT scales much better with the number of data. Std-MT offers the smallest computational burden, but has poor performance with a low number of data.}
    \label{fig:time_4}
\end{figure}

\section{Conclusions}
\review{
In this paper we have proposed a novel solution for the MTGP problem that, compared to previous formulations, eliminates the need to validate any model hyperparameters and dramatically reduces the number of parameters to be learnt. 
Similarly to other existing models, this proposal assumes that an intertask and a noise covariances exist. The novelty lies in the parameter inference, which is solved through the factorization of the joint MT likelihood into a product of conditional one-output GPs. Once these parameters are learnt, with either a hierarchical or an approximate approach, a recursive algorithm can be used to recover the MT intertask and noise covariances.
Experimental results show an accurate estimation of the MT intertask and noise matrices, which translates into an improved error performance. At the same time, we have integrated the model with standard GP toolboxes, showing that it is computationally competitive with the state of the art.}




\clearpage
\newpage





\section*{Acknowledgments and Disclosure of Funding}
\label{sec:acknowledgments}
We thank Dr. Miguel L\'azaro-Gredilla and Gustau Camps-Valls for their thorough review of the paper and fruitful discussions. This paper is part of the project PID2020-115363RB-I00 funded by MCIN/AEI/10.13039/
50110001103, the National Science Foundation EPSCoR Cooperative Agreement OIA-1757207, and the King Felipe VI Endowed Chair.
\appendix
\bibliographystyle{ieeetr}
\bibliography{MultitaskGP}

\end{document}